\documentclass[11pt]{article}

\usepackage[final]{acl}

\usepackage{times}
\usepackage{latexsym}
\usepackage[T1]{fontenc}
\usepackage[utf8]{inputenc}
\usepackage{microtype}
\usepackage{inconsolata}

\usepackage{booktabs}
\usepackage{graphicx}
\usepackage{wrapfig}
\usepackage{subcaption}
\usepackage{multirow}
\usepackage{enumitem}
\usepackage{listings}
\usepackage{amsmath,amssymb,amsfonts,amsthm,bm}
\usepackage{algorithm}
\usepackage{algorithmic}
\usepackage[frozencache,cachedir=minted-cache]{minted}
\usepackage{tcolorbox}
\tcbuselibrary{breakable,skins}
\usepackage[capitalize,nameinlink,noabbrev]{cleveref}

\definecolor{darkblue}{rgb}{0,0,0.5}
\hypersetup{colorlinks=true,citecolor=darkblue,linkcolor=darkblue,urlcolor=darkblue}

\newtheorem{definition}{Definition}
\newtheorem{assumption}{Assumption}
\newtheorem{proposition}{Proposition}

\crefname{definition}{definition}{definitions}
\Crefname{definition}{Definition}{Definitions}
\crefname{assumption}{assumption}{assumptions}
\Crefname{assumption}{Assumption}{Assumptions}
\crefname{proposition}{proposition}{propositions}
\Crefname{proposition}{Proposition}{Propositions}

\providecommand{\linenumbers}{}
\providecommand{\nolinenumbers}{}

\title{Two-Stage Reinforcement Learning for Sound and Adversarial Test Generation in Code LLMs}

\author{
Jiacheng Xu\textsuperscript{1,2},
Wentao Zhang\textsuperscript{1},
Zhiyi Lyu\textsuperscript{1},
Fuxiang Zhang\textsuperscript{1,2}, \\
\bfseries Chaojie Wang\textsuperscript{2},
Yang Liu\textsuperscript{2},
Bo An\textsuperscript{1} \\
\textsuperscript{1}Nanyang Technological University, Singapore \\
\textsuperscript{2}Skywork AI \\
\texttt{jiacheng005@e.ntu.edu.sg}, \texttt{boan@ntu.edu.sg}
}

\begin{document}
\maketitle

\begin{abstract}

Reinforcement learning (RL) has substantially advanced code generation with large language models (LLMs) through executable feedback. The feedback for coding problems mainly comes from specific test cases, where high-quality test cases are often scarce since they should be both sound and discriminative.
We thus turn to study the auto-generation of test cases using the learned model. We find this is naturally an adversarial RL problem: the model is expected to generate effective test cases as counterexamples, depending on the solver's current failure modes.
We propose Test Cases Scaling (TCS), a two-stage RL framework for effective test generation. Both stages train a test generator from a rolling policy-aligned buffer: Stage~1 generates tests consistent with the reference solution, and Stage~2 restricts the buffer to current failure modes and learns counterexample tests.
Across TACO and LiveCodeBench, TCS improves both pass@1 and inference-time answer selection according to generated tests.
We find the learned test generator also enables effective selection among other LLM outputs.

\end{abstract}

\section{Introduction}

\begin{figure}[t]
  \centering
  \includegraphics[width=\columnwidth]{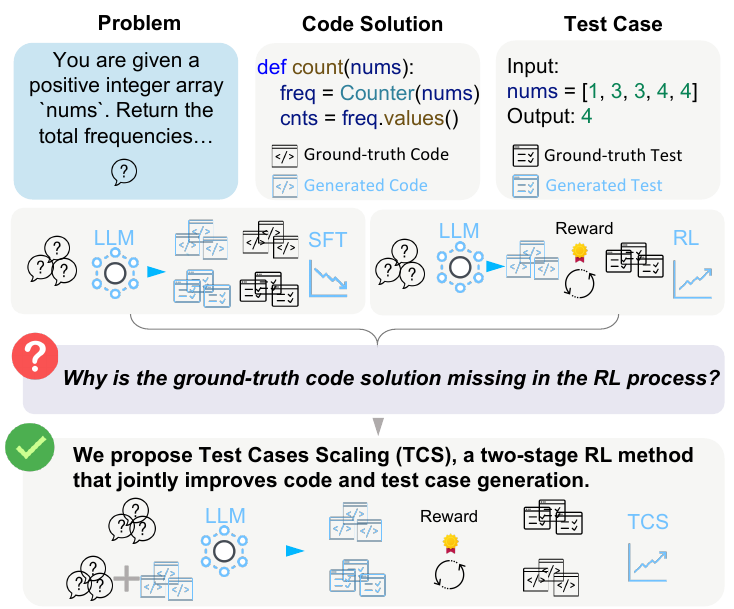}
  \caption{
    TCS enables adversarial test case generation using ground-truth solutions.
  }
  \label{fig:tcs_intro}
\end{figure}

Code generation is one of the most consequential applications of large language models (LLMs), improving developer productivity and lowering barriers for non-experts~\citep{CodeGenerationSurvey}. Modern code LLMs have moved far beyond heuristic systems~\citep{ClassicCodeGeneration,CodeLLAMA,qwencoder,CodeGenerationSurvey}, and reinforcement learning (RL) has become a particularly effective post-training tool because candidate programs can be verified by execution on test cases, often outperforming purely supervised tuning based on costly demonstrations~\citep{PPOCoder,coderl,deepcoder2025}.

As code models improve, the bottleneck shifts from producing plausible solutions to verifying which candidate is actually correct. This makes inference-time \emph{self-verification} attractive: generate tests, execute candidate programs, and select the program with the strongest empirical support. For code tasks, this is appealing because it relies on executable evidence rather than a generic scalar reward. But test-based selection helps only when generated tests are reliable and targeted. Unsound tests can mis-rank candidates, while weak tests fail to expose subtle bugs. In our experiments, self-generated tests help substantially, yet curated public tests can still outperform them when used alone, showing that the challenge is not to generate more tests, but better ones.

This challenge explains why RL is useful here beyond supervised imitation. Test generation is a multi-solution, utility-driven objective: many tests can be correct, as judged by execution rather than exact target matching. The objective is non-stationary because informative tests depend on the solver's current failure modes. Whereas code generation can optimize against a fixed test suite, test generation must produce tests that are \emph{sound} for a reference solution and \emph{discriminative} against plausible incorrect code. Optimizing only for soundness yields trivial tests, while counterexample rewards are sparse and unstable from scratch, making reward design particularly important~\citep{rewardodin,rewardhacking}.

We therefore study an execution-verifiable post-training setting in which a ground-truth solution is available to validate generated tests. This lets us separate two objectives that are easy to conflate: \emph{soundness control} through ground-truth verification and \emph{candidate-conditioned adversariality} against the model's evolving failure modes. This setting provides a clean test bed for understanding what makes learned test generation useful. Based on this view, we propose \textbf{Test Cases Scaling (TCS)}, a two-stage RL framework that jointly trains code generation and test generation with stage-specific rewards and a rolling policy-aligned buffer. Stage~1 learns tests that agree with the ground-truth solution, while Stage~2 learns candidate-conditioned counterexample tests that pass the reference solution but fail plausible incorrect programs. At inference time, the learned verifier generates tests for candidate programs, and we select the candidate with the highest pass-count across pooled tests.

Experiments on TACO and LiveCodeBench show consistent gains in both training-time and inference-time performance. The paper makes three main contributions:
\begin{itemize}
  \item We argue that effective test generation for code LLMs requires both \emph{soundness control} and \emph{candidate-conditioned adversariality}, and formulate this view in an execution-verifiable post-training setting where generated tests can be checked against a ground-truth solution.
  \item We propose \textbf{Test Cases Scaling (TCS)}, a two-stage RL framework that instantiates these principles: Stage~1 learns ground-truth-verified tests, and Stage~2 learns candidate-conditioned counterexample tests via stage-specific rewards and a policy-aligned buffer.\looseness=-1
  \item We provide theoretical and empirical evidence for test-based inference-time scaling. We derive an exponential reliability bound for pass-count selection under self-generated tests (Sec.~\ref{sec:theory}), and experiments on TACO and LiveCodeBench show improvements over joint SFT, code-only RL, and test-only RL. The learned verifier can also improve the selection of outputs from strong external LLMs.
\end{itemize}

\section{Background}

\subsection{Reinforcement Learning}
\label{sec:background_rl}

Reinforcement learning (RL) fine-tunes a policy $\pi_\theta$ to maximize task reward. We use Group Relative Policy Optimization (GRPO)~\citep{grpo}, which avoids a separate value model by using the average reward of multiple outputs from the same prompt as a baseline. For each prompt $x$, GRPO samples a group of $G$ outputs $\{y_i\}_{i=1}^G$ and computes group-relative advantages:
\begin{equation}
\label{eq:GRPO}
\begin{aligned}
\mathcal{J}(\theta)
&= \mathbb{E}_{x,\mathbf{y}}\Bigg[
\frac{1}{G}\sum_{i=1}^{G}
\min\!\Big(\rho_i A_i,\; \bar\rho_i A_i\Big)\\
&\qquad - \beta\, D_{\mathrm{KL}}(\pi_\theta \,\|\, \pi_{\mathrm{ref}})\Bigg].
\end{aligned}
\end{equation}
where the expectation abbreviates $x\sim\mathcal{D}$ and $\mathbf{y}=(y_1,\ldots,y_G)\sim\pi_{\theta_{\mathrm{old}}}^{G}(\cdot\mid x)$.
Here $\rho_i \triangleq \frac{\pi_{\theta}(y_i\mid x)}{\pi_{\theta_{\mathrm{old}}}(y_i\mid x)}$ denotes the importance sampling ratio,
$\bar\rho_i \triangleq \operatorname{clip}(\rho_i,\,1-\epsilon,\,1+\epsilon)$,
$\epsilon$ is the clipping range, $\beta$ controls the strength of KL regularization, and $D_{\mathrm{KL}}(\cdot \| \cdot)$ is the KL-divergence.
The advantage $A_i$ is defined as
\[
A_i \triangleq \frac{r_i - \operatorname{mean}(r_1,\dots,r_G)}{\operatorname{std}(r_1,\dots,r_G)},
\]
where $r_i$ is the reward on response $y_i$.

\begin{figure*}
  \centering
  \includegraphics[width=0.99\linewidth]{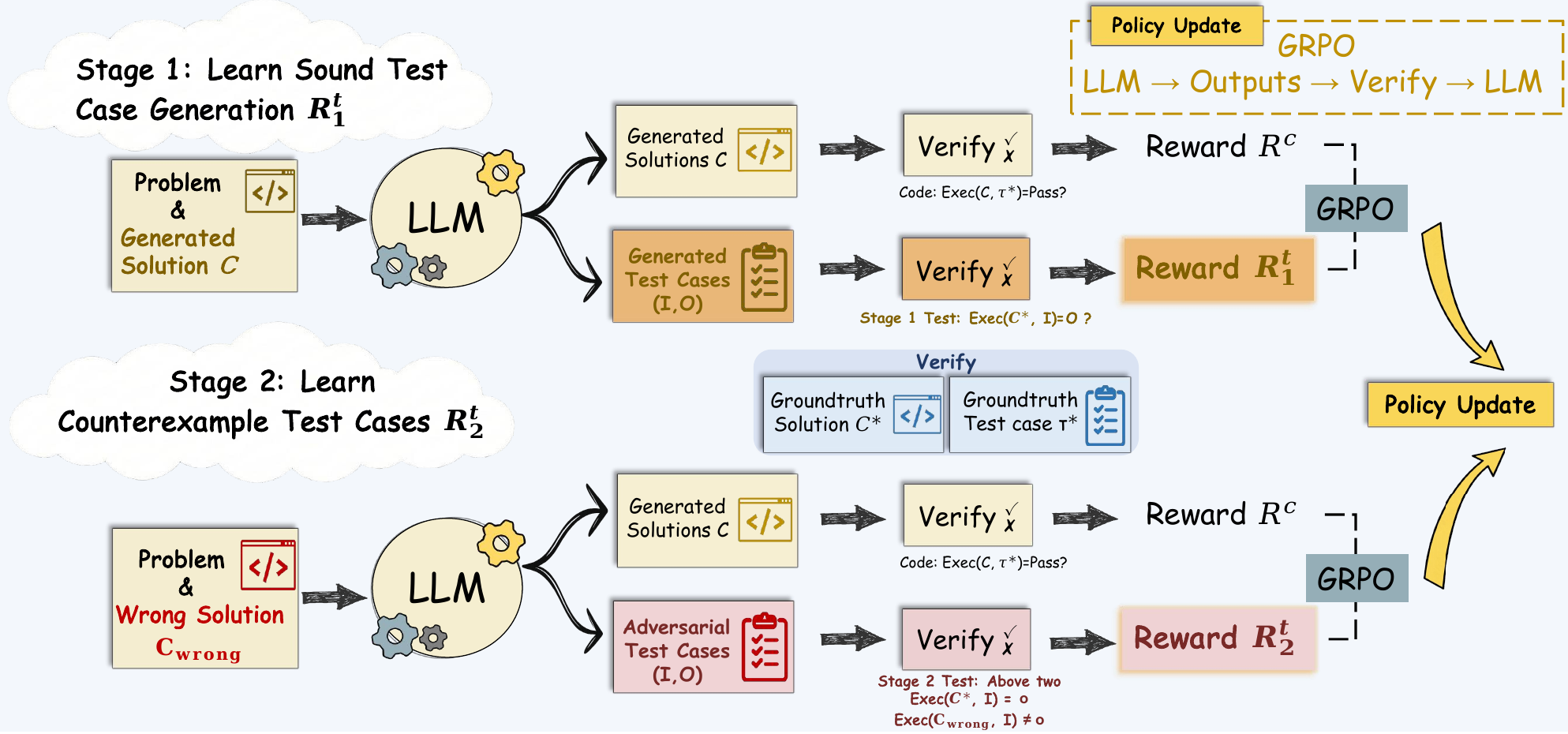}
  \caption{Overview of the two-stage training process in Test Cases Scaling (TCS).}
  \label{fig:tcs_overview}
\end{figure*}

\subsection{Code and Test Case Generation}
\label{sec:background_codegen}

The task of code generation is to produce a program \(C\) from a natural-language problem description \(P\) that satisfies the specification. Because many problems admit multiple correct implementations, candidate programs are verified against test suites \(\mathcal{T} = \{(I_i, O_i)\}_{i=1}^m\). In RL for code generation, a standard reward function $R^c$ evaluates whether a generated program $C$ passes an assumed ideal test suite $\mathcal{T}^*$~\citep{yu2025dapo}:
\begin{equation}
  \label{eq:codegen}
\begin{aligned}
R^c(C, \mathcal{T}^*) =
\begin{cases}
1, & \begin{aligned}[t]
     &\text{if } \forall (I_i, O_i) \in \mathcal{T}^*:\\
     &\mathrm{Exec}(C, I_i) = O_i
     \end{aligned} \\
0, & \text{otherwise.}
\end{cases}
\end{aligned}
\end{equation}
Thus, test quality is central to both training and evaluation. A useful test suite should be (i) \emph{sound} for the reference solution and (ii) sufficiently \emph{discriminative} to expose failures of plausible incorrect implementations. Generating tests that satisfy both properties is precisely the challenge we target in TCS.
This mirrors competitive programming practice, where constructing tests that break seemingly correct solutions is itself a core skill rather than a mere evaluation artifact.

\section{Test Cases Scaling (TCS)}

\subsection{Problem Formulation}

Each training instance consists of a problem description $P$, a ground-truth solution $C^*$, and a test suite $\mathcal{T}=\{(I_i,O_i)\}_{i=1}^m$. This setup assigns two distinct roles to the same LLM:

\begin{itemize}
    \item \textbf{Solver}: Given a problem description $P$, the solver aims to generate code $C$ that passes all test cases in the test suite $\mathcal{T}$, i.e., $\forall (I_i, O_i) \in \mathcal{T}: \mathrm{Exec}(C, I_i) = O_i$.
    \item \textbf{Verifier}: Given a problem description $P$ and a candidate program $C$, the verifier aims to generate test cases $(I_v,O_v)$ that are \emph{sound} under $C^*$ (i.e., $\mathrm{Exec}(C^*,I_v)=O_v$) and \emph{discriminative} for incorrect candidates (i.e., $\mathrm{Exec}(C,I_v)\neq O_v$ when $C$ is wrong). \looseness=-1
\end{itemize}
Unlike RL pipelines that optimize only the solver, TCS explicitly trains the same model as a verifier. The ground-truth solution provides a clean signal for \emph{soundness}, while the candidate program $C$ defines the \emph{adversarial} target. This shared-policy setup matters: the verifier is not an external checker, but a capability of the same model whose candidates it later evaluates. Verifier training is therefore part of code understanding: to produce sound counterexamples, the model must reason jointly about specifications, outputs, and bug patterns. This helps both during training, where test generation sharpens reasoning about corner cases, and during inference, where the verifier ranks candidates via self-generated tests (Sec.~\ref{TTS}).

\subsection{RL Training}

RL is well suited to this setting because execution provides direct utility feedback for a multi-solution objective. But reward design is crucial, and test generation is harder than code generation because a useful test must be both \emph{sound} for the reference solution and \emph{adversarial} against plausible incorrect code. Optimizing both at once is brittle: rewards that ignore soundness invite reward hacking, while rewards that demand full counterexamples from the start are too sparse to learn from. TCS therefore uses a staged verifier curriculum: Stage~1 learns soundness with dense feedback, and Stage~2 learns candidate-conditioned counterexamples against incorrect solutions.

We jointly train code generation and test generation (Figure~\ref{fig:tcs_overview}) with a single shared policy by mixing prompts from the code dataset $\mathcal{D}$ and verifier prompts constructed online from a policy-aligned buffer $\mathcal{B}$. We update the policy using GRPO (Eq.~\eqref{eq:GRPO}; Appendix Algorithm~\ref{alg:TCS}) with $R^c$ for solver rollouts and stage-specific verifier rewards $R^t_1$ (Eq.~\eqref{eq:stage1}) / $R^t_2$ (Eq.~\eqref{eq:stage2}) for verifier rollouts. Generated tests train the verifier and support inference-time selection, while solver rollouts are scored only by dataset tests. This joint optimization couples the solver and verifier: solver rollouts populate $\mathcal{B}$ with the policy's current behavior, and verifier rollouts learn to produce tests that are first reliable and then increasingly targeted to those evolving failure modes. Unlike fixed offline supervision, this online RL loop lets the verifier track the solver's moving error distribution.

\subsubsection{Policy-Aligned Buffer}
\label{sec:policy_aligned_buffer}

To learn candidate-conditioned tests, verifier prompts must reflect the solver's \emph{current} failure modes rather than a static offline collection of code. We therefore dynamically construct test-generation prompts from solver outputs produced online during training. Outputs that meet stage-specific criteria are collected into a rolling buffer $\mathcal{B}$, which serves as the reservoir for verifier instances. To maintain diversity and alignment with the current policy, we retain only items from the most recent $T_b$ training steps. Even in Stage~1, where the reward depends only on soundness, conditioning on current code helps the verifier learn candidate-specific contexts and keeps the prompt distribution consistent with Stage~2. The full test-generation prompt template is provided in Appendix~\ref{app:prompt_templates}.

\subsubsection{Stage 1 Training (Soundness)}

Stage~1 controls \emph{soundness}. For a ground-truth solution $C^*$ and a generated test case $(I_g, O_g)$, we assign reward $1$ if $\mathrm{Exec}(C^*, I_g)=O_g$ and $0$ otherwise. To prevent exact reuse of example tests from the prompt, we additionally require $(I_g,O_g)\notin\mathcal{T}_{\text{example}}$, yielding $R^t_1$:

\begin{equation}
  \label{eq:stage1}
  R_1^t(I_g, O_g) =
  \begin{cases}
  1, & \begin{aligned}[t]
       &\text{if }\mathrm{Exec}(C^*, I_g)=O_g \\
       &\text{and }(I_g, O_g)\notin \mathcal{T}_{\text{example}}
       \end{aligned} \\
  0, & \text{otherwise.}
  \end{cases}
  \end{equation}
where $\mathcal{T}_{\text{example}}$ denotes the example test cases in the problem description. This filter is deliberately conservative: it blocks exact copying, while the main guardrail remains execution-based verification against $C^*$.

During Stage~1, $\mathcal{B}$ admits executable solver outputs (e.g., no syntax errors or timeouts), exposing the verifier to diverse, policy-aligned code contexts without requiring strong adversarial targeting. Stage~1 ends when the verifier reaches a predefined success threshold (Appendix~\ref{sec:implementation_details}), after which we transition to Stage~2. Intuitively, Stage~1 reduces the soundness error of generated tests before optimizing for harder counterexamples.

\subsubsection{Stage 2 Training (Counterexample)}

While $R^t_1$ promotes soundness, it can still encourage trivial tests with little discriminative power. At the Stage~1$\rightarrow$Stage~2 transition, we clear $\mathcal{B}$ and thereafter admit only executable but incorrect candidates, defined as programs that fail at least one dataset test case in $\mathcal{T}$. Stage~2 then trains the verifier to generate \emph{candidate-conditioned counterexamples}: tests that pass $C^*$ but fail an incorrect candidate $C_{\text{wrong}}$ sampled from $\mathcal{B}$ (Sec.~\ref{sec:policy_aligned_buffer}). We define the adversarial reward $R^t_2$ as:

\begin{equation}
  \label{eq:stage2}
\begin{aligned}
  &R_2^t(I_g, O_g, C^{*}, C_{\text{wrong}})\\
  &\quad =
  \begin{cases}
  1, & \begin{aligned}[t]
       &\text{if }\mathrm{Exec}(C^*, I_g)=O_g \\
       &\text{and }\mathrm{Exec}(C_{\text{wrong}}, I_g)\neq O_g \\
       &\text{and }(I_g, O_g)\notin \mathcal{T}_{\text{example}}
       \end{aligned} \\
  0, & \text{otherwise.}
  \end{cases}
\end{aligned}
  \end{equation}

Despite being well motivated, $R^t_2$ can be extremely sparse early in training, especially for models with weak test-generation ability. For models like R1-Distill-Qwen-1.5B, directly optimizing $R^t_2$ yields near-zero rewards for long periods because producing a counterexample test requires both (i) generating a sound $(I_g,O_g)$ and (ii) targeting a non-trivial failure mode of $C_{\text{wrong}}$. This sparsity motivates the Stage~1$\rightarrow$Stage~2 curriculum above.

Because $R^t_2$ explicitly conditions on $C_{\text{wrong}}$, Stage~2 incentivizes targeted counterexamples rather than generic corner cases. In this sense, Stage~1 and Stage~2 instantiate the two principles highlighted in the introduction: Stage~1 controls soundness through ground-truth verification, while Stage~2 adds candidate-conditioned adversariality.

\subsection{Inference-time Scaling}\label{TTS}

Conventional inference-time scaling often samples multiple candidates and uses an external reward model to rerank. For code generation, we instead scale by generating and executing self-generated tests: we use the same test generation mechanism as in training to construct prompts conditioned on $(P,C_i)$, and then select the candidate that best satisfies the resulting tests. This selection rule is only useful when the pooled tests are sufficiently sound and sufficiently discriminative, which is exactly what TCS is designed to improve.
Given a problem, we sample $N$ candidate code solutions $\{C_i\}_{i=1}^N$. For each $C_i$, we generate $M$ test cases conditioned on $(P,C_i)$, yielding a pooled set of $K=N \times M$ self-generated tests. We execute each $C_i$ on the pooled tests and select the candidate with the highest pass-count.
Formally, our selection criterion can be expressed as:
\begin{equation}
\label{eq:inference_select}
\begin{aligned}
C_{\text{selected}}
&= \underset{C_i \in \{C_1, C_2, \ldots, C_N\}}{\arg\max}\; S(C_i),\\
S(C_i)
&= \sum_{j=1}^{N \times M}
\mathbb{I}[\mathrm{Exec}(C_i, I_j) = O_j].
\end{aligned}
\end{equation}
where $S(C_i)$ is the pass-count for the $i$-th candidate code solution, $\{(I_j, O_j)\}_{j=1}^{N \times M}$ denotes the pooled self-generated test cases, and $\mathbb{I}$ is the indicator function that equals $1$ when the execution matches the expected output and $0$ otherwise.

\subsection{Theoretical Analysis}
\label{sec:theory}
Eq.~\eqref{eq:inference_select} selects the candidate with the largest pass-count under the pooled self-generated tests. We formalize when this selection becomes reliable as the number of tests increases.

Let $\mathcal{C}=\{C_1,\dots,C_N\}$ be the candidate set, and let $\mathcal{C}^-\subset\mathcal{C}$ denote the set of incorrect candidates. Assume that $\mathcal{C}$ contains at least one correct candidate, i.e., $\mathcal{C}\setminus\mathcal{C}^-\neq\emptyset$. If $\mathcal{C}^-=\emptyset$, incorrect selection is impossible and the claim is trivial. We henceforth consider the nontrivial case $\mathcal{C}^-\neq\emptyset$. For each candidate $C_i$, we independently sample $M$ tests from the conditional test generator $G(\cdot\mid P,C_i)$, yielding a stratified pooled test set $\mathcal{T}=\{(I_{i,m},O_{i,m})\}_{i=1,m=1}^{N,M}$ of size $K=N\times M$. Throughout the analysis, generated test inputs are assumed to satisfy the input constraints of $P$. We use the induced uniform pooled-mixture distribution, which first samples $i\sim\mathrm{Unif}(\{1,\dots,N\})$ and then samples from $G(\cdot\mid P,C_i)$, only to define the aggregate quantities $\alpha$ and $\delta$ below.

\begin{definition}[Soundness error and counterexample rate]
\label[definition]{def:alpha_delta}
Let $(I,O)$ be a random test drawn from the pooled-mixture distribution above. Define the soundness error
\[
\alpha \triangleq \Pr[\mathrm{Exec}(C^*,I)\neq O],
\]
and for any incorrect candidate $C\in\mathcal{C}^-$ define the counterexample rate
\[
\begin{aligned}
\delta(C) &\triangleq
\Pr\!\left[
\begin{aligned}
&\mathrm{Exec}(C^*,I)=O\\
&\wedge\ \mathrm{Exec}(C,I)\neq O
\end{aligned}
\right],\\
\delta &\triangleq \min_{C\in\mathcal{C}^-}\delta(C).
\end{aligned}
\]
\end{definition}

$\alpha$ measures how often the test generator assigns an incorrect label to $C^*$, while $\delta$ measures how often a test makes an incorrect candidate fail while $C^*$ passes.

\begin{assumption}[Independent stratified sampling and net-discriminativeness]
\label[assumption]{ass:net_disc}
Conditioned on the fixed candidate set $\mathcal{C}$, the stratified tests $\{(I_{i,m},O_{i,m})\}_{i=1,m=1}^{N,M}$ are independent, and $\delta>\alpha$ under the induced pooled-mixture distribution.
\end{assumption}

\begin{proposition}[Exponential reliability of pass-count selection]
\label[proposition]{prop:exp_reliability}
Under \Cref{ass:net_disc}, the inference-time rule in Eq.~\eqref{eq:inference_select} selects an incorrect candidate with probability at most
\[
\begin{aligned}
&\Pr\!\big[
C_{\text{selected}} \in \mathcal{C}^-
\big]\\
&\quad \le (N-1)
\exp\!\left(-\frac{K(\delta-\alpha)^2}{2}\right).
\end{aligned}
\]
\end{proposition}
\noindent Intuitively, each additional test case changes the pass-count gap between a correct and incorrect candidate by at most $\pm 1$. When tests are net-discriminative ($\delta>\alpha$), the expected gap grows linearly with $K$ and concentrates by Hoeffding/Chernoff, yielding an exponentially small mis-selection probability. The complete proof is provided in Appendix~\ref{app:theory}.

\paragraph{Interpretation and connection to two-stage training.}
The bound in \cref{prop:exp_reliability} identifies two control knobs for inference-time scaling: the number of tests $K$ and the margin $(\delta-\alpha)$. Stage~1 (Eq.~\eqref{eq:stage1}) is designed to reduce $\alpha$ by rewarding tests consistent with $C^*$, while Stage~2 (Eq.~\eqref{eq:stage2}) is designed to increase $\delta$ by rewarding counterexample tests that fail incorrect solutions while passing $C^*$. This motivates the decoupled ablations in Appendix Table~\ref{tab:joint_vs_decoupled_main}.

\begin{table*}[t]
    \centering
  \begingroup
  \setlength{\tabcolsep}{5pt}
  \setlength{\aboverulesep}{0.35ex}
  \setlength{\belowrulesep}{0.45ex}
  \renewcommand{\arraystretch}{0.92}
    \begin{tabular*}{0.99\textwidth}{@{\extracolsep{\fill}}lccccc@{}}
\toprule
& \multicolumn{2}{c}{TACO} & \multicolumn{2}{c}{LiveCodeBench} & \\
\cmidrule(lr){2-3} \cmidrule(lr){4-5}
Model & w/o pub & w/ pub & w/o pub & w/ pub & Avg. \\
\midrule
\multicolumn{6}{c}{\textbf{DeepSeek-R1-Distill-Qwen-1.5B}} \\
\midrule
Base Model & \multicolumn{2}{c}{5.63} & \multicolumn{2}{c}{14.38} & 10.01 \\
\hspace{1em}+ Reward Model & 12.70 & 14.60 & 23.30 & 30.11 & 20.18 \\
\hspace{1em}+ Self-Generated Test Cases & 5.78 & 13.91 & 22.00 & 30.09 & 17.95 \\
\midrule
SFT Model & \multicolumn{2}{c}{9.43} & \multicolumn{2}{c}{17.21} & 13.32 \\
\hspace{1em}+ Reward Model & 14.32 & 16.78 & 23.67 & 32.45 & 21.81 \\
\hspace{1em}+ Self-Generated Test Cases & 15.12 & 16.23 & 24.33 & 31.89 & 21.89 \\
\midrule
\textbf{RL using TCS Training} & \multicolumn{2}{c}{12.31} &\multicolumn{2}{c}{20.63} & 16.47 \\
\hspace{1em}+ Reward Model & 15.66 & 23.17 & 24.01 & 37.99 & 25.21 \\
\textbf{\hspace{1em}+ Self-Generated Test Cases} & \textbf{20.52} & \textbf{25.18} & \textbf{27.47} & \textbf{38.90} & \textbf{28.02} \\
\midrule
\multicolumn{6}{c}{\textbf{DeepSeek-R1-Distill-Qwen-7B}} \\
\midrule
Base Model & \multicolumn{2}{c}{14.36} & \multicolumn{2}{c}{28.56} & 21.46 \\
\hspace{1em}+ Reward Model & 24.87 & 26.98 & 42.65 & 46.24 & 35.19 \\
\hspace{1em}+ Self-Generated Test Cases & 18.67 & 26.02 & 43.01 & 46.15 & 33.46 \\
\midrule
SFT Model & \multicolumn{2}{c}{18.92} & \multicolumn{2}{c}{32.14} & 25.53 \\
\hspace{1em}+ Reward Model & 27.45 & 29.67 & 43.23 & 48.91 & 37.32 \\
\hspace{1em}+ Self-Generated Test Cases & 27.32 & 29.78 & 44.67 & 47.82 & 37.40 \\
\midrule
\textbf{RL using TCS Training} & \multicolumn{2}{c}{24.09} & \multicolumn{2}{c}{37.03} & 30.56 \\
\hspace{1em}+ Reward Model & 31.11 & 37.67 & 43.01 & 53.40 & 41.30 \\
  \textbf{\hspace{1em}+ Self-Generated Test Cases} & \textbf{35.35} & \textbf{39.40} & \textbf{48.79} & \textbf{54.75} & \textbf{44.57} \\
  \bottomrule
  \end{tabular*}
  \endgroup
  \caption{Performance on TACO and LiveCodeBench.}
  \label{main_results}
\end{table*}

\section{Experiments}

\subsection{Experimental Setup}

Our method assumes an execution-verifiable setting: each training example contains a problem description, a ground-truth solution, and reliable test cases, so generated tests can be checked during post-training. We use TACO~\citep{TACO} for training and evaluation, filtering its 25,433 problems down to 6,318 curated instances with adequate test coverage and at least one Python solution that passes all cases. We evaluate on the TACO validation set (1,000 problems) and LiveCodeBench~\citep{LiveCodeBench}. The Stage~1$\rightarrow$Stage~2 switch is a fixed hard transition aligned with roughly $0.75$ training-batch test accuracy (200 steps for 1.5B, 40 for 7B), and inference-time selection uses $M=1$ generated test per candidate. Full details are deferred to Appendix~\ref{sec:implementation_details}.

\subsection{Main Results}

We evaluate DeepSeek-R1-Distill-Qwen models (1.5B and 7B) in three settings: the base model, an offline joint code--test SFT baseline, and TCS. The SFT baseline uses the same training problems as TCS with R1-Distill-Qwen-32B sample generation, following Sol-Ver~\citep{solververify}: code supervision comes from teacher samples that pass the reference tests, and test supervision comes from candidate-conditioned, ground-truth-verified cases that expose the paired incorrect code sample. Thus, the baseline already receives execution-verified adversarial supervision, but only through a fixed offline dataset rather than TCS's policy-aligned online updates. In TCS, we enter Stage~2 once test-generation accuracy exceeds a threshold.

Table~\ref{main_results} reports pass@1 (16 samples on TACO and 32 on LiveCodeBench) and Best-of-$N$ selected by either reward-model ranking or self-generated tests ($N=16$ for TACO, $N=32$ for LiveCodeBench, and $M=1$ generated test per candidate). We use InternLM2-7B-reward~\citep{rewardmodel} and the CodeT~\citep{CodeT} pipeline for these two selection rules, both with and without public test cases.

Table~\ref{main_results} summarizes both training-time and inference-time performance. TCS improves pass@1 for both 1.5B and 7B models relative to the base and joint-SFT baselines, showing that verifier training helps not only reranking but also the solver itself. More importantly, TCS yields the strongest gains when self-generated tests are used for selection, indicating that the learned verifier becomes materially more useful after training. Since the joint-SFT baseline already receives candidate-conditioned, ground-truth-verified adversarial supervision offline, this gap reflects the value of online RL rather than merely better labels.

The table clarifies why \emph{soundness control} matters. For the base model, self-generated tests are often weaker than reward-model ranking, reflecting unreliable test generation. SFT narrows this gap but still leaves test-based selection inconsistent. TCS changes this pattern: with matched backbones, self-generated tests become competitive with or stronger than reward-model ranking, especially without public test cases, when selection relies almost entirely on the learned verifier.

On LiveCodeBench, the 7B base model improves from a pass@1 of 28.56 to 43.01 with self-generated test selection at BoN-32. Under the same protocol, curated public tests alone reach 45.99, while combining public and self-generated tests works best at 46.15. Self-generated tests therefore provide substantial value without automatically surpassing curated tests; this is precisely where soundness control matters. TCS addresses this with Stage~1 for reliability and Stage~2 for candidate-conditioned counterexamples. Appendix Figure~\ref{fig:test_time_scaling_7B_public_case} extends this comparison across sample counts, with and without public test cases, and shows the same pattern: self-generated tests help most when they complement reliable filtering signals rather than replace them blindly.

Importantly, these gains do not come from just any external test generator. Under the same inference-time budget, tests generated by CodeRM-8B~\citep{coderm} yield weaker filtering performance than both reward-model ranking and our self-generated tests (Appendix~\ref{app:coderm_baseline}).

\paragraph{Joint vs.\ decoupled training.}
To separate the effect of \emph{joint} training from generic RL gains, we compare TCS with two \emph{decoupled} baselines on the same R1-Distill-Qwen-1.5B backbone: code-only RL (Code-RL) and test-only RL (Test-RL). Appendix Table~\ref{tab:joint_vs_decoupled_main} reports pass@1 and no-public-test selection results.

These results separate solver-side from verifier-side RL. \textbf{Code-RL} improves pass@1 over the base model, but adds little under ``+TC,'' suggesting better solutions without a much better verifier. In contrast, \textbf{Test-RL} yields a stronger verifier: although its direct pass@1 remains modest, its self-generated tests produce much larger gains with ``+TC,'' consistent with a higher counterexample rate $\delta$. \textbf{Joint TCS} training achieves both the best training-time performance and the largest inference-time gains under self-generated test selection, consistent with jointly reducing soundness error and enlarging $(\delta-\alpha)$ in Sec.~\ref{sec:theory}. Filtering TCS-generated code with Test-RL tests still underperforms TCS self-verification, consistent with a synergistic self-play effect: a stronger solver creates harder failure modes for the verifier. Appendix Table~\ref{tab:rl_training_with_only_code_generation_task} likewise shows that code-only RL does not reproduce the same test-based selection gains. This rules out the simpler explanation that TCS works only because RL improves the solver: explicit verifier training matters independently.

\subsection{Inference-Time Scaling Comparison}\label{exp:TTS}

\begin{figure}[t]
  \centering
  \includegraphics[width=\columnwidth]{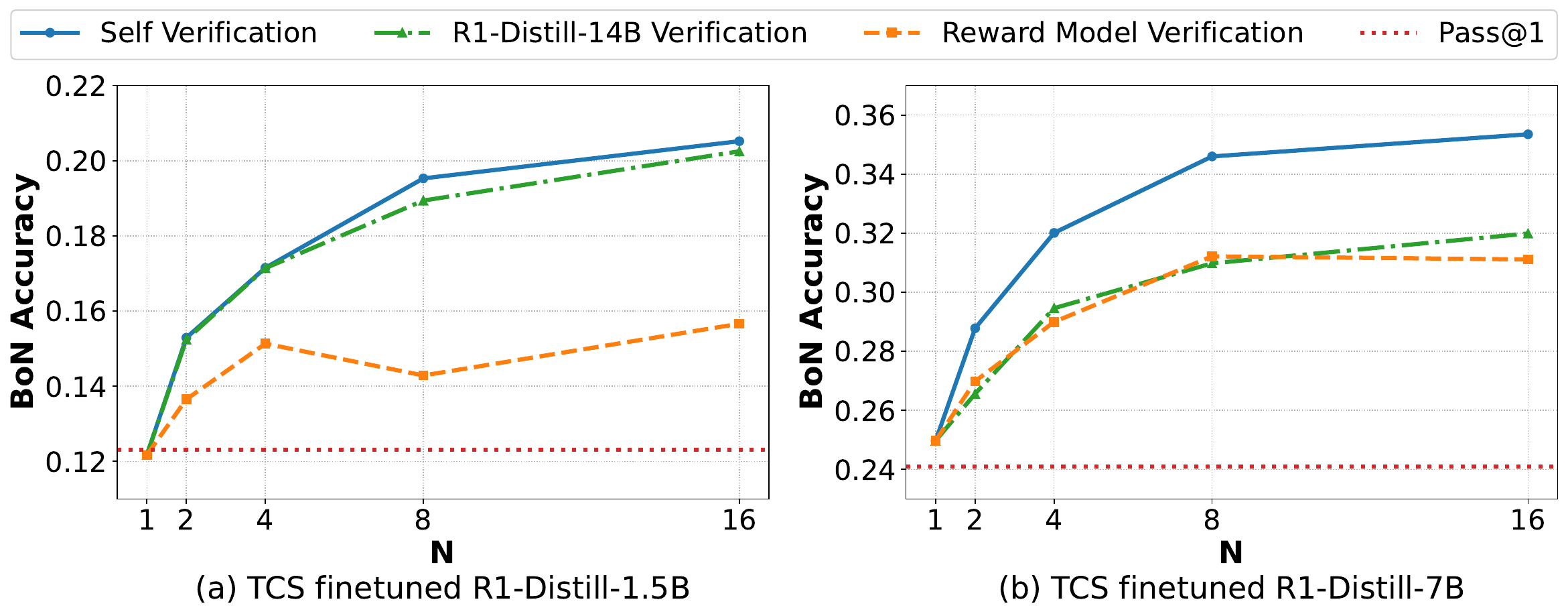}
  \caption{Comparison of different inference-time scaling methods on TACO.}
  \label{fig:test_time_scaling}
\end{figure}

Inference-time computation often improves answer quality by sampling multiple candidates and ranking them with an auxiliary signal. For code generation, one natural signal is execution on self-generated tests. We use DeepSeek-R1-Distill-Qwen-14B as a strong baseline and apply the same selection rule as in Section~\ref{TTS}. Figure~\ref{fig:test_time_scaling} shows that self-generated test selection yields consistent gains as $N$ increases. Notably, the TCS-fine-tuned 1.5B model can outperform the much larger 14B baseline under this rule. Reward-model selection is stronger than naive self-generation for weaker models, but becomes less stable as $N$ grows, suggesting sensitivity to out-of-distribution candidates. In contrast, TCS yields a more robust test-based scaling signal, consistent with Sec.~\ref{sec:theory}: additional tests help only when their distribution keeps $\alpha$ low while providing enough $\delta$.

\begin{figure*}[t]
  \centering
  \begin{subfigure}[b]{0.32\textwidth}
      \centering
      \includegraphics[width=\textwidth]{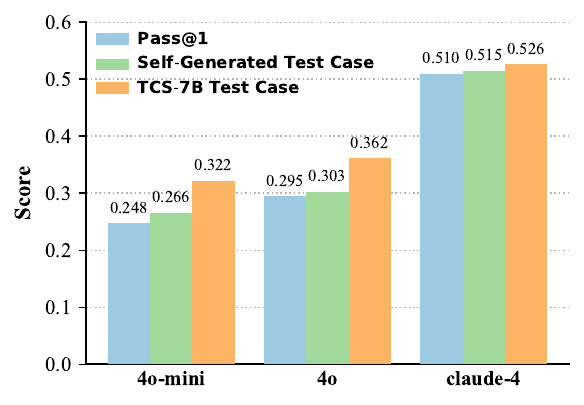}
  \end{subfigure}
  \hfill
  \begin{subfigure}[b]{0.32\textwidth}
      \centering
      \includegraphics[width=\textwidth]{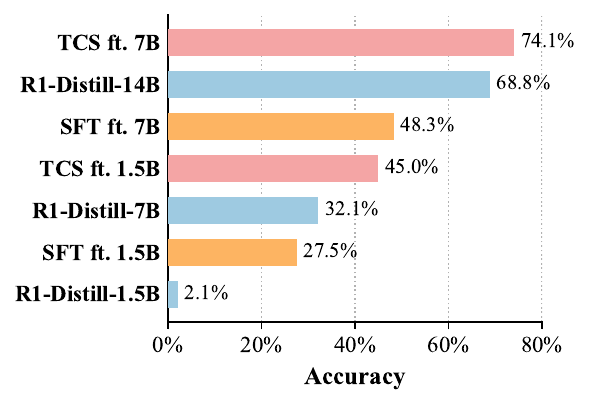}
  \end{subfigure}
  \hfill
  \begin{subfigure}[b]{0.32\textwidth}
      \centering
      \includegraphics[width=\textwidth]{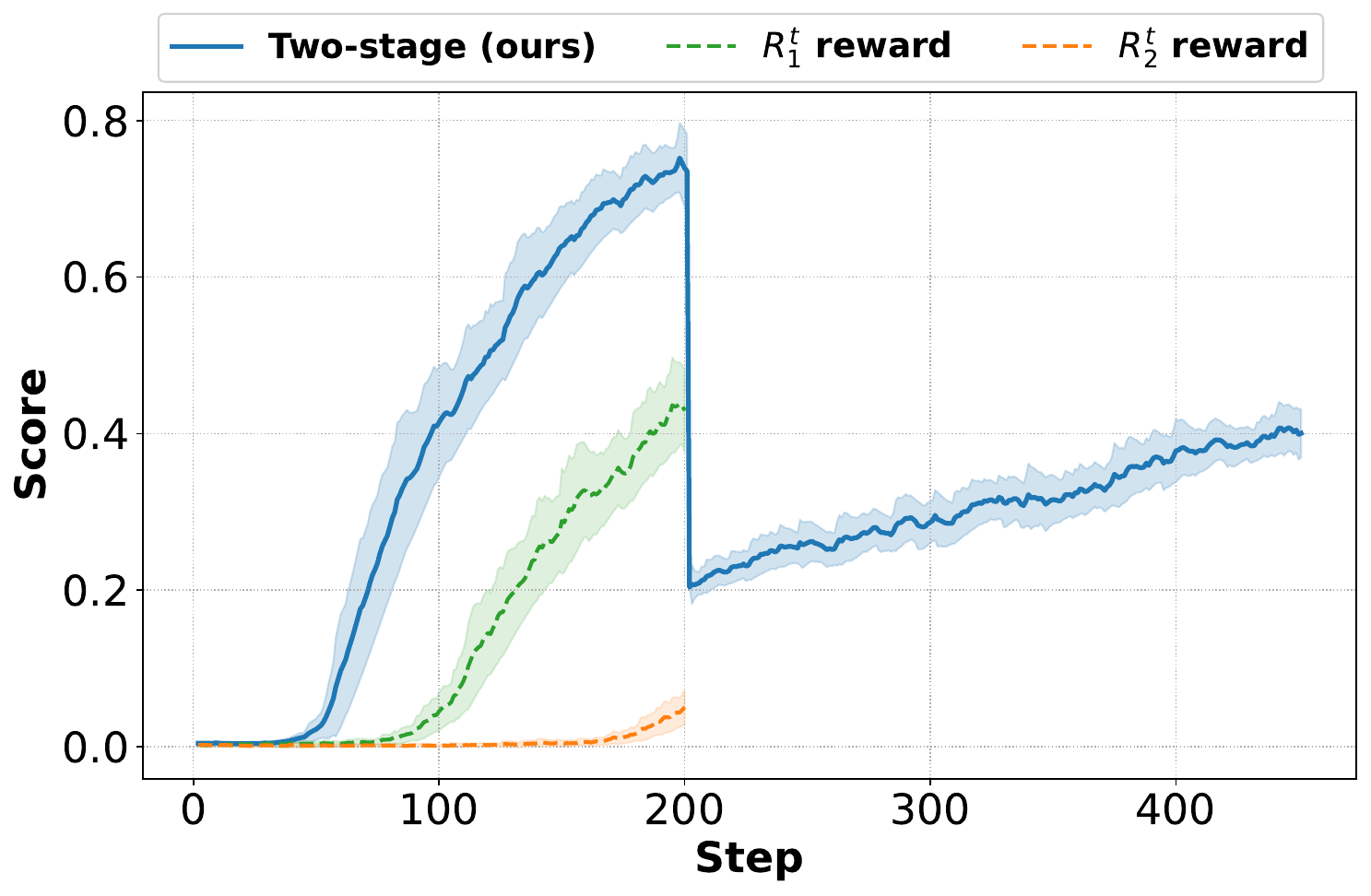}
  \end{subfigure}
  \caption{(a) Pass@1 (blue) and BoN (N=8) (green/orange) performance selected by self-generated or TCS-7B-generated test cases for strong external models in LiveCodeBench. (b) Test case output prediction accuracy. The abbreviation ``ft.'' denotes finetuned models. (c) Scores under different training rewards.}
  \label{fig:comparison_test_cases}
\end{figure*}

\subsection{Test Case Effectiveness}

To better understand why TCS helps, we evaluate generated tests along two axes. First, we measure \emph{practical filtering power}: do the generated tests improve Best-of-$N$ selection for strong external models? Second, we measure \emph{output consistency} using the official LiveCodeBench Test Output Prediction task: given the problem and a gold-test input, the model predicts the corresponding output, which is scored against the verified reference. We use this prediction accuracy as a proxy for test reliability and soundness, rather than as a complete measure of adversariality by itself.

Figure~\ref{fig:comparison_test_cases}(a) shows that tests generated by TCS-7B yield larger downstream selection gains on LiveCodeBench than those from strong external models, indicating stronger practical filtering power. Figure~\ref{fig:comparison_test_cases}(b) shows that TCS models achieve higher test-output prediction accuracy, which we interpret as a proxy for stronger soundness and consistency. We therefore use output-prediction accuracy to characterize reliability, and downstream selection gains to characterize usefulness for candidate filtering. Together, these results support the claim that TCS improves both the reliability and practical value of generated tests. Appendix Figures~\ref{fig:adversarial_test_case_example} and~\ref{fig:adversarial_test_case_success} further illustrate the adversarial tests learned by TCS, including cases that eliminate incorrect solutions even after they pass all available public tests.

\subsection{Effectiveness of Two-Stage Reinforcement Learning}

We hypothesize that Stage~1 and Stage~2 play distinct roles: Stage~1 mainly improves soundness, while Stage~2 mainly improves adversariality. Under the same computational budget, Stage~1-only training yields only modest test-based scaling gains (Appendix Table~\ref{tab:stage_reward}), showing that soundness alone is insufficient as $N$ grows. By contrast, the full two-stage reward produces clearer gains, especially when public and self-generated tests are combined.

Figure~\ref{fig:comparison_test_cases}(c) plots batch-mean test-generation rewards over training steps and shows that directly optimizing $R^t_2$ from the outset yields sparse rewards and ineffective counterexample learning for many steps. Stage~1 therefore establishes a sufficiently sound verifier before Stage~2, enabling more effective Stage~2 training.

\section{Related Work}

Post-training for code LLMs has primarily focused on improving the solver, with large gains from SFT and RL on code generation tasks~\citep{qwencoder,guo2024deepseek,jiang2024training,fan2025fait}. More recent work has also explored unit-test generation, verifier learning, and solver-verifier co-training~\citep{solververify,li2025codei,pytest,CURE}. Against this backdrop, we focus on a narrower question: in an execution-verifiable setting, what properties make learned test generation useful for inference-time selection? Accordingly, we study matched-backbone, matched-budget evidence for soundness control and candidate-conditioned adversariality, operationalized through a staged RL objective and a policy-aligned buffer.

Recent inference-time scaling approaches improve quality by sampling multiple candidates and selecting among them using an auxiliary signal~\citep{inferencetime,wu2024inference}. For code, prior work has used reward models, interpreters, search, and self-generated tests to rerank candidate programs~\citep{rewardmodel,inferencetimerewardmodel,shinn2023reflexion,CodeT,TestChain,light2024scattered,yu2024outcome,tang2024code}. Our focus differs in three ways: we train the verifier itself, optimize verifier utility with RL rather than only imitating a fixed offline test distribution, and analyze reliability through the soundness error $\alpha$ and counterexample rate $\delta$.

\section{Conclusion}

We argued that effective test generation for code LLMs requires soundness control and candidate-conditioned adversariality, and that RL is useful here because test generation is multi-solution, execution-defined, and policy-dependent. Across TACO and LiveCodeBench, Test Cases Scaling (TCS) improves both code generation and self-verification; theory and ablations support the same picture: inference-time scaling strengthens as soundness error decreases and counterexample rate increases.
The experiments also suggest a practical lesson: self-generated tests are most useful when reliability is controlled, and they work best when they complement rather than simply replace curated public tests. The consistent gap over the offline joint-SFT baseline further indicates that policy-aligned online RL matters beyond merely exposing the model to fixed adversarial supervision.
Appendix Table~\ref{tab:detailed_results_taco} further shows that these gains persist across TACO difficulty levels and become more pronounced on harder subsets.
The transfer gains on strong external models further suggest that the learned verifier captures broadly useful failure modes rather than only self-play artifacts.

\section*{Limitations}

The primary limitation of our method is that, during inference-time test case generation, it currently produces only a single test case per inference call. Generating multiple test cases in a single inference would be more efficient, but this is hindered by the challenge of defining an appropriate reward function: intuitive metrics such as counting correct cases or measuring accuracy rates are susceptible to reward hacking. Investigating reinforcement learning strategies that enable simultaneous generation of multiple test cases thus remains a promising direction. Additionally, our current approach uses a hard switch between training stages. Exploring soft switching by dynamically adjusting the proportion of the two types of test case reward functions may yield improved results. We did not pursue this primarily due to the high resource demands of RL for LLMs, and therefore prioritized a direct and reliable method. Our framework also assumes access to a ground-truth solution \(C^*\) during post-training to verify generated tests. Although \(C^*\) is not required at inference, this assumption limits training in settings without verified solutions. Extending TCS to model-based or consistency-based verification is an important direction for future work~\citep{selfverify,selfconsistency}. Regarding societal impact, while enhanced LLM coding capabilities can improve productivity, they may also increase risks of misuse, such as in interviews and competitions.

\bibliography{custom}

\clearpage
\appendix
\section{Implementation Details}
\label{sec:implementation_details}

\subsection{Data}

The TACO dataset meets our requirements for the execution-verifiable training setting studied in this paper: each example provides a problem, a comprehensive test suite, and submitted solutions (which may not always be correct). The original dataset contains 25,433 problems. To ensure accurate evaluation, we first filter out problems with fewer than 50 test cases, leaving 10,605 problems. We then remove problems without any submitted solutions, as these are necessary for evaluating generated test cases, resulting in 7,918 problems. Next, we verify the submitted solutions against the test cases to ensure the problems are verifiable, since some may be of unverifiable types (e.g., multi-turn input). We retain only those problems for which at least one solution passes all test cases, yielding a final training set of 6,318 problems. 
The processed TACO training split is publicly available directly on Hugging Face: \href{https://huggingface.co/datasets/XiaoBanni/TACO-Train}{TACO-Train}.

We use two widely adopted benchmarks in code-LLM evaluation: TACO~\citep{TACO} and LiveCodeBench~\citep{LiveCodeBench}. Following standard practice, we include LiveCodeBench problems from August 2024 to February 2025 in our evaluation~\citep{skyworkor1}. Since TACO serves as the post-training source while the LiveCodeBench window begins in August 2024, there is a chronological separation between the post-training corpus and the LiveCodeBench evaluation set.

For TACO evaluation, which lacks public test cases, the first evaluation test case is designated as public. When public test cases are available, candidate solutions are filtered to include only those passing these tests before applying inference-time selection methods.

\subsection{Experimental Configuration}

We use verl~\citep{verl} as our RL framework and adopt the default experimental configuration except for the following modifications:

\begin{itemize}[leftmargin=*, itemsep=2pt, parsep=0pt]
    \item batch size: 128
    \item PPO mini-batch size: 64
    \item GRPO group size: 16
    \item temperature (for both training and evaluation): 0.8
    \item maximum response length: 8192
\end{itemize}

We do not use the KL loss, and we use the entropy loss to sustain the model's entropy.

For DeepSeek-R1-Distill-Qwen-1.5B, the number of training steps is 200 for stage 1 and 250 for stage 2; the code generation baseline is 450 steps. For DeepSeek-R1-Distill-Qwen-7B, the number of training steps is 40 for stage 1 and 160 for stage 2; the code generation baseline is 200 steps. In both cases, the transition from stage 1 to stage 2 is determined by test-case-generation accuracy measured on the training batches. In practice, we use a hard switch once this quantity reaches approximately $0.75$, which corresponds to step 200 for the 1.5B model and step 40 for the 7B model. We choose the threshold of $0.75$ because it lies near the knee of the Stage~1 learning curve.

Code-RL and TCS are trained for the same number of RL training steps at each model size: 450 steps for 1.5B and 200 steps for 7B.

For inference-time scaling, we set $M=1$ in the main experiments, meaning that each sampled code candidate is paired with exactly one generated test case per inference call.

Both of the following TCS-finetuned checkpoints are publicly available on Hugging Face: \href{https://huggingface.co/XiaoBanni/TCS-1.5B}{TCS-1.5B} and \href{https://huggingface.co/XiaoBanni/TCS-7B}{TCS-7B}.

We use TRL~\citep{vonwerra2022trl} as our SFT framework. The SFT baseline uses the same training problems as TCS, but replaces online RL with fixed offline supervision. We use DeepSeek-R1-Distill-Qwen-32B to generate 32 samples for each problem. For the code generation task, we retain the samples that pass all the reference test cases. For the test case generation task, we construct candidate-conditioned examples from sampled code and retain generated tests that (i) are verified by the ground-truth solution and (ii) expose the paired incorrect code sample. We then train on the mixed code--test SFT dataset with the default TRL configuration for 3 epochs.

For closed-source model experiments in Figure~\ref{fig:comparison_test_cases}(a), we use the API to generate 8 samples for each problem and use the default API configuration throughout.

Our code and evaluation pipeline are publicly available in the \href{https://github.com/xiaobanni/tcs}{TCS GitHub repository}.

\subsection{Computing Resources}
\label{sec:computing_resources}

The computing resources used at each stage, based on the NVIDIA H100 80GB, are summarized in Table~\ref{tab:computing_resources} in terms of approximate GPU hours consumed.

\begin{table*}[t]
\centering
\begin{tabular}{lccc}
\toprule
\textbf{Model} & \textbf{Training} & \textbf{Eval on LiveCodeBench} & \textbf{Eval on TACO} \\
\midrule
R1-Distill-Qwen-1.5B & 720 & 10 & 20 \\
R1-Distill-Qwen-7B & 960 & 16 & 32 \\
\bottomrule
\end{tabular}
\caption{GPU hours on NVIDIA H100 80GB for training and single-run evaluation.}
\label{tab:computing_resources}
\end{table*}

Note that the times reported here refer to the training or evaluation time for a single run of each model, not the total time across all experiments. The rollout response number for each question is 32 for LiveCodeBench and 16 for TACO. The evaluation time includes both code generation and test case generation. Using decoded tokens as a hardware-independent cost proxy, one generated test uses 22--34\% as many tokens as one code candidate in the Table~\ref{main_results} evaluation responses (22--24\% for 1.5B and 32--34\% for 7B), and independent test-generation calls can run in parallel.

\subsection{Training Pseudocode}

\begin{algorithm}[h]
  \caption{Test Cases Scaling}
  \label{alg:TCS}
  \begin{algorithmic}[1]
  \STATE \textbf{Input:} Dataset $\mathcal{D}$, initial policy $\pi_{\theta}$, buffer size $T_b$, group size $G$, total training steps $T$
  \STATE \textbf{Initialize:} Policy-aligned buffer $\mathcal{B}\leftarrow \emptyset$
  \FOR{$t=1, \dots, T$}
      \STATE Sample a data batch from the joint dataset $\mathcal{D} \cup \mathcal{B}$
      \FOR{input prompt $x$ in the batch}
          \STATE Generate a group of responses $\{y_i\}_{i=1}^{G}$ from $\pi_{\theta}$
          \IF{the input $x$ is from $\mathcal{D}$}
              \STATE $\triangleright$ Code generation problem
              \STATE Compute rewards $r^c$ for the group $\{y_i\}_{i=1}^{G}$ according to \cref{eq:codegen}
              \STATE Append to $\mathcal{B}$ all $(x, y_i)$ pairs from the group meeting the current buffer criteria.
          \ELSE
              \STATE $\triangleright$ Test case generation problem
              \STATE Compute rewards $r^t_1$ or $r^t_2$ for the group $\{y_i\}_{i=1}^{G}$ according to \cref{eq:stage1} or \cref{eq:stage2}
              \STATE Remove $x$ from $\mathcal{B}$
          \ENDIF
      \ENDFOR
      \STATE Update the model $\pi_\theta$ according to \cref{eq:GRPO}
      \STATE Remove data samples from $\mathcal{B}$ that were collected before $t - T_b$ steps
  \ENDFOR
  \end{algorithmic}
\end{algorithm}

\subsection{Adversarial Reward Computation}
\label{sec:stage2_details}

Each generated test case is evaluated only against its paired program $C$. For Stage~2, we do not aggregate performance across multiple incorrect codes $C_{\text{wrong}}$, since the objective is to construct a targeted counterexample for a specific error pattern. As long as a test case successfully exposes the flaw in the paired $C_{\text{wrong}}$, it is treated as a \emph{sound} adversarial example (reward $1$); it is not required to act as a universal counterexample that simultaneously fails other incorrect implementations. Regarding robustness, we distinguish between buffer admission and reward computation: during buffer construction, we filter out syntax errors to ensure basic executability, while at reward time we explicitly treat robustness failures as successes. Concretely, if a generated test case is sound for the ground-truth solution $C^*$ (i.e., $C^*$ executes without error and produces the expected output) but causes the targeted $C_{\text{wrong}}$ to raise a runtime error or timeout, we still assign reward $1$, encouraging the model to propose corner cases that reveal both logical and robustness defects (e.g., infinite loops or unhandled exceptions during execution).

\subsection{Stage-1 Prompt Conditioning and Example-Copy Filter}

During Stage~1, the verifier prompt already includes a candidate solution even though $R_1^t$ depends only on soundness under $C^*$. This is intentional: it keeps the verifier's prompt format aligned with Stage~2 and teaches the model to reason in candidate-conditioned contexts before the stricter adversarial reward is introduced.

For the non-copy condition in $R_1^t$, we compare the generated input-output pair against the example pairs in $\mathcal{T}_{\text{example}}$ and assign zero reward to exact matches. This check is intentionally conservative. It is designed to prevent trivial reuse of prompt examples, not to guarantee difficulty by itself. The stronger safeguards are execution-based soundness in Stage~1 and discriminative pressure from $R_2^t$ in Stage~2.
\section{Additional Theoretical Analysis and Proofs}
\label{app:theory}

This appendix provides a formal statement and proof of the exponential reliability bound used to justify the inference-time selection rule in Eq.~\eqref{eq:inference_select}, and discusses the assumptions under which the bound matches our pooled test generation procedure described below.

\subsection{Formal setup}
Fix a problem $P$ and a candidate set of $N$ code solutions $\mathcal{C}=\{C_1,\dots,C_N\}$. Let $\mathcal{C}^-\subset\mathcal{C}$ denote the set of incorrect candidates. If $\mathcal{C}^-=\emptyset$, incorrect selection is impossible and the claim is trivial; below we therefore consider $\mathcal{C}^-\neq\emptyset$. Fix a ground-truth solution $C^*$, which is available during training and used to verify generated tests, and fix a correct solution $C^\dagger\in\mathcal{C}\setminus\mathcal{C}^-$ that agrees with $C^*$ on all valid inputs. Throughout the analysis, every generated test input is assumed to satisfy the input constraints of $P$.

At inference time, for each candidate $C_i$, we sample $M$ tests from a conditional test generator $G(\cdot\mid P,C_i)$ and pool them. Concretely, denote the pooled tests by
\[
\mathcal{T}=\{(I_{i,m},O_{i,m})\}_{i=1,m=1}^{N,M},
\qquad K \triangleq N M.
\]
We score any program $C$ by the pooled pass-count
\[
S(C)\triangleq \sum_{i=1}^{N}\sum_{m=1}^{M} \mathbb{I}[\mathrm{Exec}(C,I_{i,m})=O_{i,m}],
\]
and select $C_{\text{selected}}\in\arg\max_{C\in\mathcal{C}} S(C)$ (ties can be broken arbitrarily; treating ties as ``potentially incorrect'' only strengthens our upper bound).

\paragraph{Execution convention.}
We view $\mathrm{Exec}(C,I)$ as returning either a concrete output or a special symbol $\bot$ indicating runtime error/timeout. In the indicator above, we interpret $\mathrm{Exec}(C,I)=O$ as \emph{exact match} of outputs; in particular, $\mathrm{Exec}(C,I)=\bot$ always counts as a failure.

\paragraph{Pooled-mixture distribution.}
Although tests are generated \emph{stratified} by candidates (exactly $M$ per $C_i$), it is convenient to define the induced pooled-mixture distribution $\mathsf{P}_{\text{pool}}$ as:
sample $i\sim \mathrm{Unif}(\{1,\dots,N\})$, then sample $(I,O)\sim G(\cdot\mid P,C_i)$.
Our analysis will only require that the $K$ pooled tests are independent (not necessarily identically distributed); the mixture viewpoint is used purely to define aggregate quantities such as $\alpha$ and $\delta$ below.

\subsection{Definitions: soundness and counterexample rate}
For a random test $(I,O)\sim \mathsf{P}_{\text{pool}}$, define the soundness error
\[
\alpha \triangleq \Pr[\mathrm{Exec}(C^*,I)\neq O].
\]
For an incorrect candidate $C\in\mathcal{C}^-$, define the counterexample rate
\[
\begin{aligned}
\delta(C) \triangleq \Pr[&
\mathrm{Exec}(C^*,I)=O\\
&{}\wedge\ \mathrm{Exec}(C,I)\neq O],
\end{aligned}
\]
and $\delta\triangleq \min_{C\in\mathcal{C}^-} \delta(C)$.

\subsection{Proof of \cref{prop:exp_reliability}}
Fix an incorrect candidate $C\in\mathcal{C}^-$. Define the score gap
\[
\begin{aligned}
D(C)&\triangleq S(C^\dagger)-S(C)
      =\sum_{i=1}^{N}\sum_{m=1}^{M} X_{i,m},\\
X_{i,m} &\triangleq
\mathbb{I}[\mathrm{Exec}(C^\dagger,I_{i,m})=O_{i,m}]\\
&\quad -\mathbb{I}[\mathrm{Exec}(C,I_{i,m})=O_{i,m}].
\end{aligned}
\]
Each $X_{i,m}\in[-1,1]$. If an incorrect candidate $C$ is selected, then $S(C)\ge S(C^\dagger)$, and hence $D(C)\le 0$. Therefore,
\[
\Pr[\text{incorrect selection}]
\le \sum_{C\in\mathcal{C}^-}\Pr[D(C)\le 0].
\]
It remains to upper bound $\Pr[D(C)\le 0]$.

Fix $i\in\{1,\dots,N\}$ and consider a single test $(I,O)\sim G(\cdot\mid P,C_i)$. Define the \emph{per-candidate} soundness error and counterexample rate:
\[
\begin{aligned}
\alpha_i &\triangleq \Pr[\mathrm{Exec}(C^*,I)\neq O \mid i],\\
\delta_i(C) &\triangleq \Pr[
\mathrm{Exec}(C^*,I)=O\\
&\qquad {}\wedge\ \mathrm{Exec}(C,I)\neq O \mid i].
\end{aligned}
\]
On the \emph{sound} event $\mathrm{Exec}(C^*,I)=O$, the input $I$ is valid by assumption, so correctness of $C^\dagger$ implies $\mathrm{Exec}(C^\dagger,I)=O$; hence $X_{i,m}=1$ whenever $\mathrm{Exec}(C,I)\neq O$, which occurs with probability $\delta_i(C)$.
The only way to obtain $X_{i,m}=-1$ is if $C$ matches the (possibly incorrect) expected output while $C^\dagger$ does not, which can only happen when $\mathrm{Exec}(C^*,I)\neq O$, an event of probability $\alpha_i$. Therefore,
\[
\mathbb{E}[X_{i,m}\mid i] \ge \delta_i(C)-\alpha_i.
\]
By independence of the pooled tests, 
\[
\begin{aligned}
\mathbb{E}[D(C)]
&= \sum_{i,m}\mathbb{E}[X_{i,m}]\\
&\ge \sum_{i,m}\big(\delta_i(C)-\alpha_i\big)\\
&= M\sum_{i=1}^{N}\big(\delta_i(C)-\alpha_i\big).
\end{aligned}
\]
By the definition of the pooled-mixture distribution, these per-candidate quantities average to the corresponding pooled quantities below:
\[
\delta(C)=\frac{1}{N}\sum_{i=1}^{N}\delta_i(C),
\qquad
\alpha=\frac{1}{N}\sum_{i=1}^{N}\alpha_i,
\]
so $\mathbb{E}[D(C)] \ge K(\delta(C)-\alpha) \ge K(\delta-\alpha)$.
Applying Hoeffding's inequality for a sum of independent bounded variables (each with range length $2$) yields
\[
\begin{aligned}
\Pr[D(C)\le 0]
&=\Pr\!\big[
D(C)-\mathbb{E}D(C)\\
&\qquad\le -\mathbb{E}D(C)\big]\\
&\le \exp\!\left(-\frac{K(\delta-\alpha)^2}{2}\right).
\end{aligned}
\]
Finally, a union bound over at most $N-1$ incorrect candidates gives
\[
\Pr[\text{incorrect selection}]
\le (N-1)e^{-K(\delta-\alpha)^2/2}.
\]

\subsection{Discussion of assumptions and failure modes}
\textbf{Candidate contains a correct solution.} The bound conditions on the event that $\mathcal{C}$ contains at least one correct $C^\dagger$. Without this, selection is necessarily incorrect.

\textbf{Why require $\delta>\alpha$?} The condition $\delta>\alpha$ ensures a positive expected score gap between a correct candidate and any incorrect competitor under the pooled test distribution. If $\delta\le \alpha$, then in the worst case unsound or inconsistent tests can overwhelm the counterexample signal, and pass-count selection may not concentrate around the correct candidate; thus no meaningful exponential guarantee is possible unless additional assumptions are imposed.

\textbf{Independence and the pooled-mixture viewpoint.} In practice, tests are generated conditionally on each candidate $C_i$, and they may exhibit dependencies due to shared decoding randomness or prompt structure. Our proof only uses that the pooled tests are independent and bounded, and defines $\alpha,\delta$ via the induced uniform mixture $\mathsf{P}_{\text{pool}}$. If strong dependencies exist, one can replace Hoeffding with concentration for martingales or an effective sample size; the qualitative dependence on $K$ and the margin $(\delta-\alpha)$ remains the same.

\textbf{Worst-case $\delta$ is conservative.} Using $\delta=\min_{C\in\mathcal{C}^-}\delta(C)$ yields a compact worst-case guarantee. Retaining the candidate-specific rates gives the sharper bound
\[
\begin{aligned}
&\Pr[\text{incorrect selection}]\\
&\quad\le \sum_{C\in\mathcal{C}^-}
\exp\!\left(-\frac{K(\delta(C)-\alpha)^2}{2}\right).
\end{aligned}
\]
An average-case bound would require an explicit distribution over candidate sets or incorrect competitors, which we do not assume here.

\section{Additional Experimental Results}

\subsection{Inference-Time Scaling Comparison}

In this section, we present additional experiments to further analyze the inference-time scaling behavior of our TCS-fine-tuned 7B model when strong public cases are provided. LiveCodeBench provides around 2.64 public cases per question and is a relatively strong filter for incorrect code.

\begin{figure*}[t]
    \centering
    \includegraphics[width=0.98\textwidth]{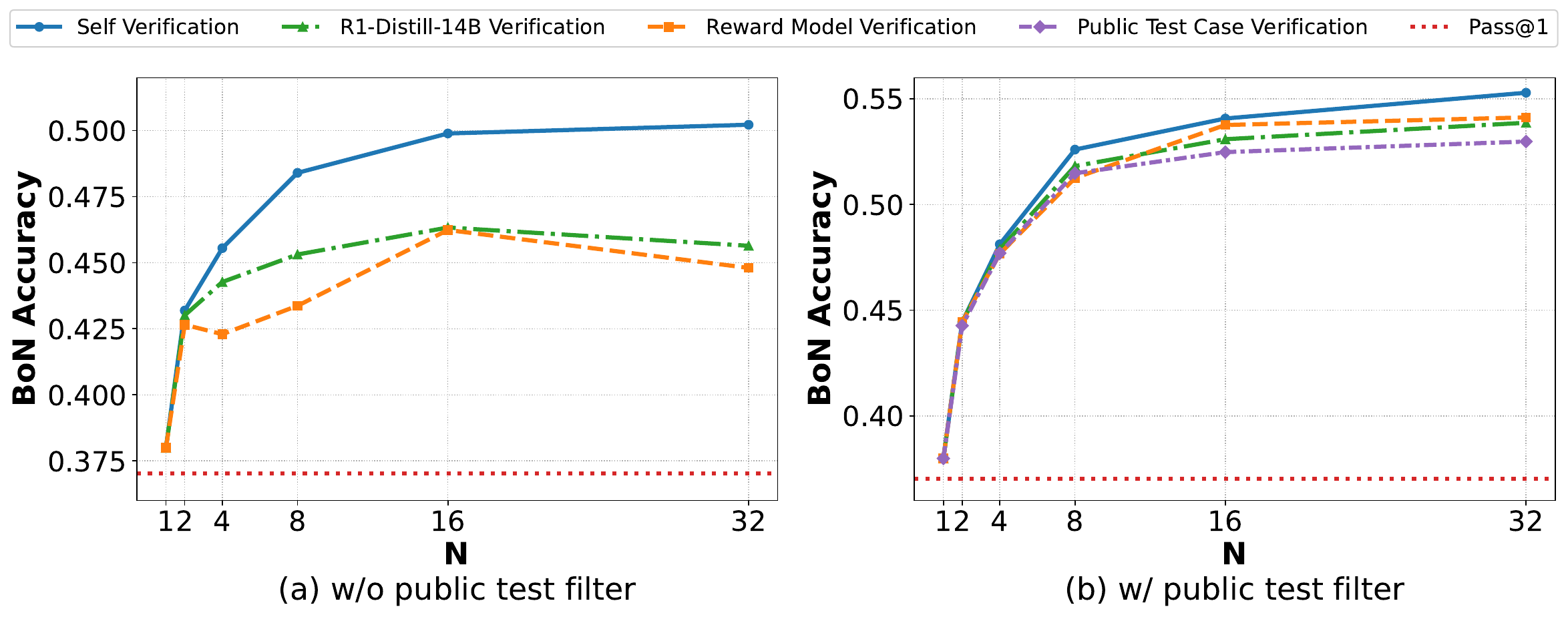}
    \caption{Comparison of different inference-time scaling methods for the TCS-fine-tuned 7B model on LiveCodeBench, with and without public cases.}
    \label{fig:test_time_scaling_7B_public_case}
\end{figure*}

Figure~\ref{fig:test_time_scaling_7B_public_case}(a) shows the inference-time scaling results without public test cases. As the number of sampled candidates increases, the baselines can degrade due to additional distractors and out-of-distribution (OOD) candidates; for reward-model selection this can reduce ranking reliability, and for test-case-based selection with an untrained test generator it can introduce noisy tests that mis-rank candidates. In contrast, the TCS-fine-tuned model maintains consistent gains as the number of candidates increases.

Figure~\ref{fig:test_time_scaling_7B_public_case}(b) shows that when strong public test cases are provided, all methods improve as the number of sampled candidates increases, but TCS retains a clear advantage. This supports the practical setting where public tests are weak or absent: the model can rely on self-generated tests for selection and can further augment any available public cases with more targeted tests.

\paragraph{Scaling the number of tests per candidate.}
We further vary $M$ while fixing $N=16$ on TACO and $N=32$ on LiveCodeBench under the no-public-test protocol. As shown in Table~\ref{tab:m_scaling}, selection performance improves monotonically from $M=1$ to $M=4$ in all four settings, providing empirical support for the dependence on $K=NM$ in Proposition~\ref{prop:exp_reliability}. Meanwhile, $M=1$ already improves pass@1 by 6.84--11.76 points, motivating its use as the default in the main experiments.

\begin{table}[t]
  \centering
  \small
  \setlength{\tabcolsep}{2.5pt}
  \begin{tabular}{llcccc}
    \toprule
    Model & Benchmark & pass@1 & $M=1$ & $M=2$ & $M=4$ \\
    \midrule
    TCS-1.5B & TACO & 12.31 & 20.52 & 21.45 & 22.13 \\
    TCS-1.5B & LCB  & 20.63 & 27.47 & 28.26 & 29.27 \\
    TCS-7B   & TACO & 24.09 & 35.35 & 36.09 & 36.43 \\
    TCS-7B   & LCB  & 37.03 & 48.79 & 49.43 & 50.53 \\
    \bottomrule
  \end{tabular}
  \caption{Inference-time scaling with multiple independently generated tests per candidate. We fix $N=16$ on TACO and $N=32$ on LiveCodeBench and use no public test cases.}
  \label{tab:m_scaling}
\end{table}

\subsection{Joint vs.\ Decoupled Training}

Table~\ref{tab:joint_vs_decoupled_main} restores the full comparison between joint TCS training and the two decoupled RL baselines. All results are reported without public test cases. Rows marked ``+TC'' apply self-generated test selection, and the final row ``+Test-RL-TC'' filters TCS-generated code with tests generated by the Test-RL model. The pattern is consistent with the discussion in the main text: Code-RL mainly improves the solver, Test-RL mainly improves the verifier, and joint TCS improves both.

\begin{table}[t]
  \centering
  \small
  \begin{tabular}{lcc}
    \toprule
    Model & TACO (w/o pub) & LCB (w/o pub) \\
    \midrule
    Base & 5.63 & 14.38 \\
    \hspace{0.8em}+TC & 5.78 & 22.00 \\
    \midrule
    Code-RL & 11.16 & 18.53 \\
    \hspace{0.8em}+TC & 11.92 & 21.05 \\
    \midrule
    Test-RL & 7.85 & 15.95 \\
    \hspace{0.8em}+TC & 15.35 & 23.26 \\
    \midrule
    \textbf{TCS} & \textbf{12.31} & \textbf{20.63} \\
    \hspace{0.8em}\textbf{+TC} & \textbf{20.52} & \textbf{27.47} \\
    \hspace{0.8em}+Test-RL-TC & 18.79 & 25.94 \\
    \bottomrule
  \end{tabular}
  \caption{Joint vs.\ decoupled training on R1-Distill-Qwen-1.5B. ``TC'' denotes selection with self-generated test cases; ``Test-RL-TC'' uses test cases generated by the Test-RL model.}
  \label{tab:joint_vs_decoupled_main}
\end{table}

\subsection{Effectiveness of Two-Stage Reinforcement Learning}

Table~\ref{tab:stage_reward} reports the full reward-ablation results referenced in the main text. Stage~1-only training yields limited scaling gains as the number of candidates grows, whereas the full two-stage reward becomes increasingly advantageous, especially when public and self-generated tests are combined. This complements Figure~\ref{fig:comparison_test_cases}(c), which shows that applying the stricter Stage~2 reward from the outset leads to sparse and unstable learning.

\begin{table}[t]
  \centering
  \scriptsize
  \setlength{\tabcolsep}{2pt}
  \begin{tabular*}{\columnwidth}{@{\extracolsep{\fill}}llccccc@{}}
  \toprule
  \textbf{Reward} & \textbf{Filter} & \textbf{N=1} & \textbf{N=2} & \textbf{N=4} & \textbf{N=8} & \textbf{N=16} \\
  \midrule
  \multirow{2}{*}{S1 only}
      & Public               & 24.23    & 29.89 & 33.00 & 34.89 & 35.83 \\
      & P+S                  & 24.23    & 29.68 & 33.40 & 35.80 & 36.70 \\
  \cmidrule(lr){1-7}
  \multirow{2}{*}{2-stage}
      & Public               & 24.09    & 28.89 & 32.37 & 33.99 & 36.26 \\
      & P+S                  & 24.09    & 29.31 & 33.82 & 37.38 & 39.40 \\
  \bottomrule
  \end{tabular*}
  \caption[Inference-time scaling comparison for the 7B model on TACO with varying reward settings.]{Inference-time scaling comparison for the 7B model on TACO with varying reward settings. Abbrev.: S1 = Stage-1 reward; 2-stage = two-stage reward; P = public test cases; S = self-generated test cases.}
  \label{tab:stage_reward}
\end{table}

\subsection{RL Training with Only the Code Generation Task}

\begin{table}[t]
    \centering
    \footnotesize
    \setlength{\tabcolsep}{2pt}
    \renewcommand{\arraystretch}{0.9}
    \begin{tabular*}{\columnwidth}{@{\extracolsep{\fill}}p{0.36\columnwidth}cccc@{}}
      \toprule
      & \multicolumn{2}{c}{TACO} & \multicolumn{2}{c}{LCB} \\
      \cmidrule(lr){2-3} \cmidrule(lr){4-5}
      Model & w/o pub & w/ pub & w/o pub & w/ pub \\
      \midrule
      \multicolumn{5}{c}{\textbf{DeepSeek-R1-Distill-Qwen-1.5B}} \\
      \midrule
      Base Model & \multicolumn{2}{c}{5.63} & \multicolumn{2}{c}{14.38} \\
      \hspace{1em}+ Reward Model & 12.70 & 14.60 & 23.30 & 30.11 \\
      \hspace{1em}+ SGTC & 5.78 & 13.91 & 22.00 & 30.09 \\
      \midrule
      RL by Code Training & \multicolumn{2}{c}{11.16} & \multicolumn{2}{c}{18.53} \\
      \hspace{1em}+ Reward Model & 15.87 & 22.54 & 20.79 & 34.50 \\
      \hspace{1em}+ SGTC & 11.92 & 21.68 & 21.05 & 33.90 \\
      \midrule
      \textbf{RL by TCS Training} & \multicolumn{2}{c}{12.31} &\multicolumn{2}{c}{20.63} \\
      \hspace{1em}+ Reward Model & 15.66 & 23.17 & 24.01 & 37.99 \\
      \textbf{\hspace{1em}+ SGTC} & \textbf{20.52} & \textbf{25.18} & \textbf{27.47} & \textbf{38.90} \\
      \bottomrule
    \end{tabular*}
  \caption{Performance on TACO and LiveCodeBench. SGTC = self-generated test case selection.}
  \label{tab:rl_training_with_only_code_generation_task}
  \end{table}

In this section, we explore whether RL training exclusively on our collected data for code generation, rather than for test case generation, can indirectly improve the model’s capability in generating test cases. The results in Table~\ref{tab:rl_training_with_only_code_generation_task} show that the accuracy gains in test case generation from RL with code training are limited. This result highlights the necessity of treating test case generation as an independent task that requires dedicated RL to achieve substantial improvements.

\subsection{Detailed Results in TACO Evaluation}

There are five difficulty levels in the TACO evaluation. In this section, we present the detailed results for each difficulty level. (TC) means using test cases generated by the model itself to choose the best solution.
Our results demonstrate that finetuning with our approach yields significant improvements in code generation performance, surpassing both the base model and the base model with self-verification. The benefits of our method become increasingly evident as task difficulty rises. Additionally, leveraging self-generated test cases delivers further inference-time scaling gains, with particularly notable improvements on more challenging tasks. These findings highlight the adaptability of our approach to difficult problems and its potential applicability to stronger models and more complex code generation tasks.
\begin{table}[H]
    \centering
    \scriptsize
    \setlength{\tabcolsep}{2pt}
    \begin{tabular*}{\columnwidth}{@{\extracolsep{\fill}}lcccccc@{}}
        \toprule
        & \textbf{EASY} & \textbf{MED} & \textbf{MED\_HARD} & \textbf{HARD} & \textbf{V\_HARD} & \textbf{TOTAL} \\
        \midrule
        \textbf{R1-7B} & 34.87 & 24.78 & 13.09 & 2.75 & 0.72 & 14.36 \\
        \textbf{R1-7B(TC)} & 43.71 & 32.32 & 17.20 & 4.00 & 1.53 & 18.67 \\
        \textbf{TCS-7B} & 46.11 & 37.53 & 24.44 & 10.56 & 6.66 & 24.09 \\
        \textbf{TCS-7B(TC)} & 58.88 & 50.55 & 34.43 & 25.31 & 12.76 & 35.35 \\
        \bottomrule
    \end{tabular*}
    \caption{Performance comparison across different difficulty levels and total scores for various models.}
    \label{tab:detailed_results_taco}
\end{table}

\subsection{CodeRM-8B Baseline Comparison}
\label{app:coderm_baseline}

We additionally compare our TCS-fine-tuned 7B model with CodeRM-8B~\citep{coderm}, a unit-test-generation model obtained by supervised fine-tuning (SFT) Llama-3.1-8B-Instruct on test cases synthesized by a larger Llama-3.1-70B-Instruct teacher. To ensure a fair comparison, we use the same inference-time budget $N$ for all methods when sampling code candidates and associated test cases. The pass@1 and Best-of-$N$ selection results on TACO and LiveCodeBench are summarized in \cref{tab:coderm_comparison}.

\begin{table}[H]
  \centering
  \footnotesize
  \setlength{\tabcolsep}{2pt}
  \renewcommand{\arraystretch}{0.9}
  \begin{tabular*}{\columnwidth}{@{\extracolsep{\fill}}p{0.42\columnwidth}cccc@{}}
    \toprule
    \multirow{2}{*}{Method} & \multicolumn{2}{c}{TACO} & \multicolumn{2}{c}{LCB} \\
    \cmidrule(lr){2-3} \cmidrule(lr){4-5}
    & w/o pub & w/ pub & w/o pub & w/ pub \\
    \midrule
    RL by TCS-7B Training & \multicolumn{2}{c}{24.09} & \multicolumn{2}{c}{37.03} \\
    \hspace{1em}+ Reward Model & 31.11 & 37.67 & 43.01 & 53.40 \\
    \hspace{1em}+ CodeRM-8B & 23.67 & 35.90 & 41.70 & 51.49 \\
    \hspace{1em}+ Self-Generated & \textbf{35.35} & \textbf{39.40} & \textbf{48.79} & \textbf{54.75} \\
    \bottomrule
  \end{tabular*}
  \caption{Comparison of different inference-time selection methods for the TCS-fine-tuned 7B model, using either a reward model, CodeRM-8B-generated tests, or self-generated tests.}
  \label{tab:coderm_comparison}
\end{table}

We observe that CodeRM-8B's test cases yield weaker filtering performance than both our reward-model baseline and our self-generated tests. We attribute this to three factors. First, the base model underlying CodeRM-8B is less capable than the TCS-7B backbone, limiting the quality of its generated tests. Second, CodeRM-8B relies purely on SFT, whereas our method uses RL; as shown in Table~\ref{main_results}, under the same base model RL consistently outperforms SFT for code-related tasks. Third, CodeRM-8B acts as an external verifier, while TCS jointly trains the solver and verifier using a policy-aligned buffer, so the test generator is explicitly optimized to expose the current failure modes of the solver. This tight alignment between policies leads to more targeted and effective bug discovery than a generic external verifier.

\section{Generated Adversarial Test Case Examples}

A correct response in training stage two is to generate a test case that the ground-truth solution passes but the targeted incorrect solution fails. Figure~\ref{fig:adversarial_test_case_example} provides an example. The generated test case has a valid input, and the correctness of its output is verified by the ground-truth code. This test case can efficiently reveal errors in the incorrect code, so it is a desired adversarial test case. The following is a snippet from the complete reasoning process that motivates this construction:

\begin{figure*}[t]
    \centering
    \includegraphics[width=0.9\textwidth]{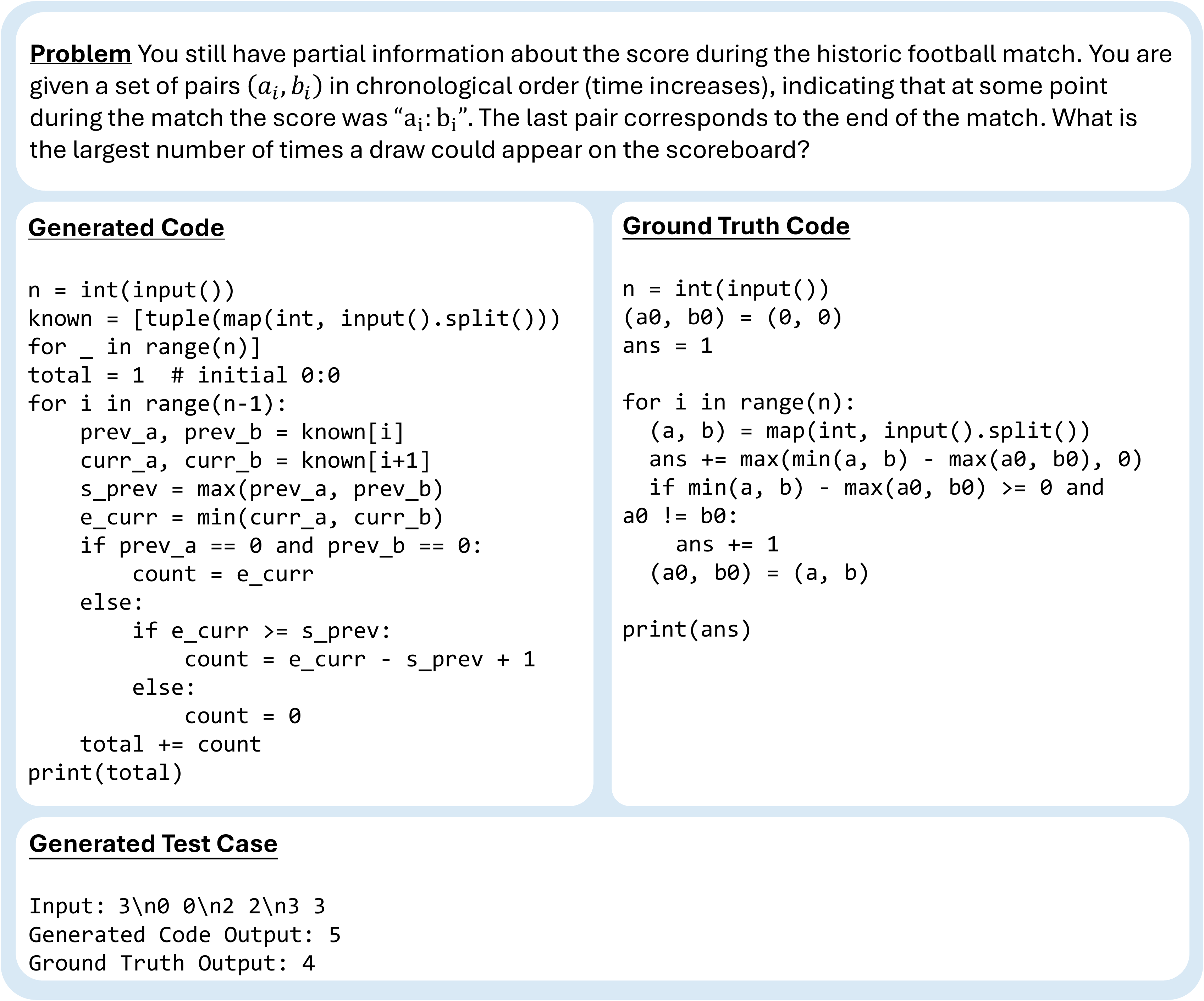}
    \caption{A generated test case that receives a reward in training stage two.}
    \label{fig:adversarial_test_case_example}
\vspace{-10pt}
\end{figure*}

\begin{quote}
``...But maybe that's incorrect. For example, if the last known draw was at x and the next known is y where y > x, but in between, maybe there were more draws possible? Or perhaps it's only considering overlapping ranges.

Wait, maybe in some cases, the code is not correctly handling multiple overlapping intervals. For example, between two known points, the number of possible draws could be more than what the code calculates.

Let me think of a test case that would expose an error in the code...''
\end{quote}

\begin{figure*}[t]
    \centering
    \includegraphics[width=0.9\textwidth]{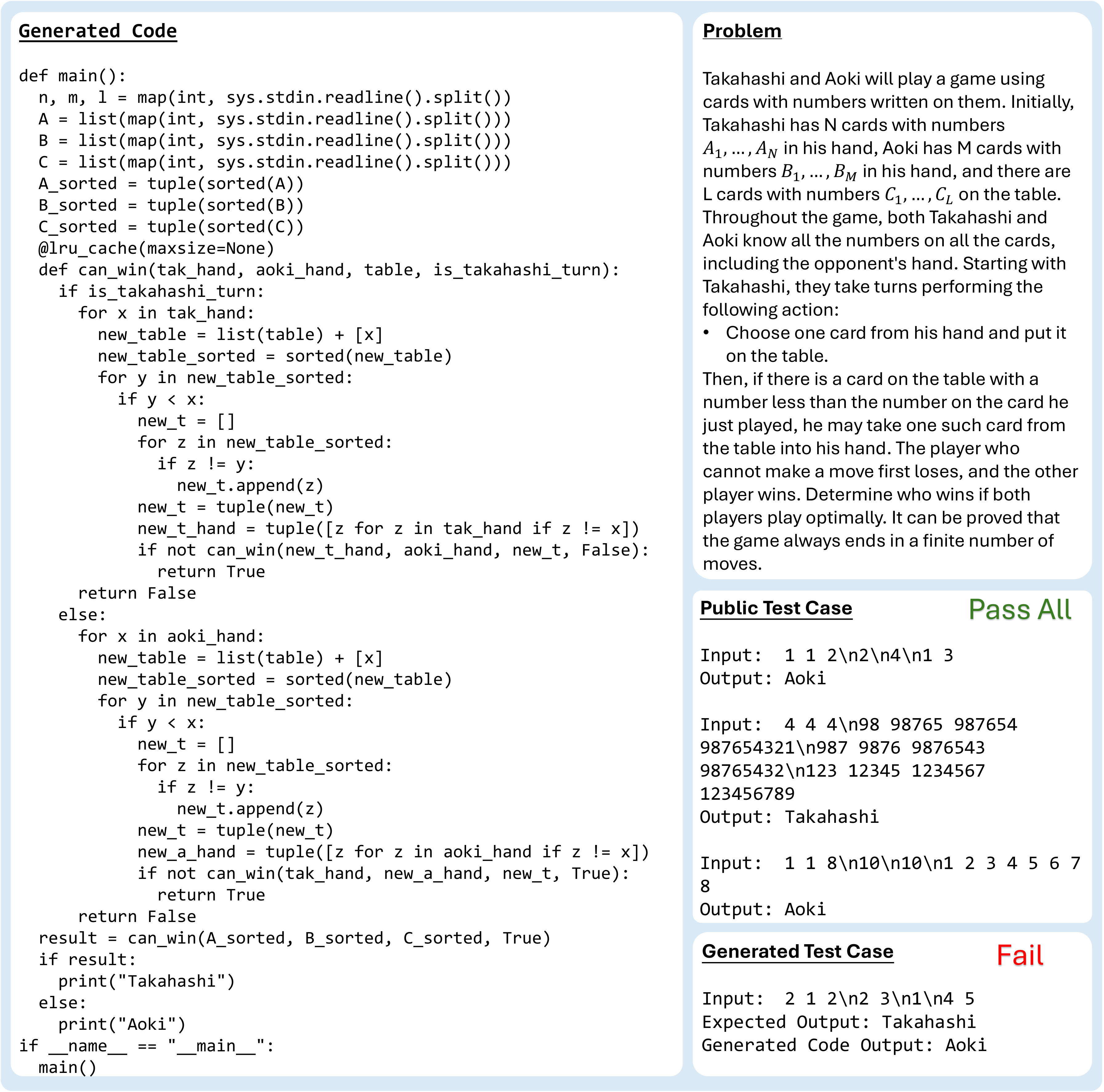}
    \caption{A generated test case that can identify an error in the generated code, even though it passes all the public test cases.}
    \label{fig:adversarial_test_case_success}
\vspace{-10pt}
\end{figure*}

To demonstrate that our TCS-fine-tuned model can generate high-quality, challenging, and adversarial test cases, we present a generated test case that filters out incorrect code that already passes all public test cases, as shown in Figure~\ref{fig:adversarial_test_case_success}. The problem is a difficult AtCoder challenge with relatively strong public test cases. During inference, we generate 32 responses, and three of them pass all public test cases. Using our generated test case, we filter out two solutions with hard-to-detect errors (one is shown in the figure), selecting the only correct answer. We provide the full reasoning process for this test case in Listing~\ref{lst:reasoning_for_test_case}.

\section{Prompt Template}
\label{app:prompt_templates}

\begin{tcolorbox}[title=Prompt Template for Code Generation,fonttitle=\bfseries]
\begin{minted}[fontsize=\small,breaklines,escapeinside=||,
  breakindent=0pt,
  breaksymbolleft={},
  breaksymbolright={}]{text}
You will be given a question (problem specification) and will generate a correct Python program that matches the specification and passes all tests.
|{\color{blue}{\{problem\}}}|
Read the inputs from stdin solve the problem and write the answer to stdout (do not directly test on the sample inputs). Enclose your code within delimiters as follows. Ensure that when the python program runs, it reads the inputs, runs the algorithm and writes output to STDOUT.
```python
# YOUR CODE HERE
\end{minted}
\end{tcolorbox}

We provide all the templates used for code generation, test case generation, and test case output prediction in this section.

\paragraph{Code Generation Prompt}
This prompt is the same as the one used in the LiveCodeBench benchmark for code generation.

\paragraph{Test Case Generation Prompt}
There are four configurable items in the test case generation prompt template. For each \textbf{problem}, we require the model to generate a test case that can reveal potential errors in a specific \textbf{code}. We also provide the format of example \textbf{test cases} for reference, but the generated test case should be different from these examples.  
To improve the diversity of the prompts, we define four \textbf{types} of test cases and randomly sample one when constructing a test case:

\begin{itemize}[leftmargin=*, itemsep=2pt, parsep=0pt]
    \item \textbf{basic}: basic test case that validates core functionality with simple, straightforward inputs.
    \item \textbf{edge}: edge case that tests boundary values and constraint limits (minimum/maximum allowed values)
    \item \textbf{corner}: corner case with unusual inputs like empty collections, single elements, or patterns that might break naive solutions
    \item \textbf{performance}: performance test with large inputs approaching the problem's limits to evaluate solution efficiency
\end{itemize}

\begin{tcolorbox}[title=Prompt Template for Test Case Generation,fonttitle=\bfseries,breakable,enhanced,
boxrule=0.8mm,
segmentation style={solid,draw=black,line width=0.8mm},
pad at break=0mm]
\begin{minted}[
  fontsize=\small,
  breaklines,
  escapeinside=||,
  breakindent=0pt,
  breaksymbolleft={},
  breaksymbolright={}
]{text}
You are an expert TEST CASE GENERATOR for programming competitions. Your ONLY task is to generate ONE TEST CASE of a specific type, NOT to solve the problem or write any implementation code.

Follow these strict guidelines:
1. DO NOT write any solution code in any programming language.
2. DO NOT attempt to fix or improve the solution.
3. FOCUS EXCLUSIVELY on generating a |{\color{blue}{\{test\_case\_type\}}}| that can reveal flaws or confirm correctness.
4. The test case you generate MUST NOT be identical to any Example Test Case provided in the problem statement.

You have been provided with:
* Problem Description:
|{\color{blue}{\{problem\}}}|
* A SOLUTION CODE THAT MAY CONTAIN LOGIC ERRORS:
```python
|{\color{blue}{\{code\}}}|
```
Your goal is to create a test case that is valid under the problem constraints and is likely to expose incorrect behavior in the provided code if such errors exist.

For the test case you create, provide:
* The test input exactly as it would be fed to a program.
* The expected output that a correct solution should produce.
* A brief explanation of what aspect this test case is verifying or how it could reveal flaws.

Ensure the output follows the *Expected Output Format* structure provided. You must enclose the output in a ```json``` block to facilitate easy extraction and processing.

Expected Output Format:
```json
|{\color{blue}{\{expected\_output\}}}|
```

You must ONLY provide ONE |{\color{blue}{\{test\_case\_type\}}}|. Ensure your test case is valid according to the problem constraints.
\end{minted}
\end{tcolorbox}

\paragraph{Test Case Output Prediction Prompt}
In this prompt, we strictly require the model to output the prediction in the expected format to ensure that the answer can be correctly extracted.

\begin{tcolorbox}[title=Prompt Template for Test Case Output Prediction,fonttitle=\bfseries]
\begin{minted}[fontsize=\small,breaklines,escapeinside=||,
  breakindent=0pt,
  breaksymbolleft={},
  breaksymbolright={}]{text}
You are an expert TEST CASE GENERATOR for programming competitions. Now you are given a test case input, and you need to infer what its output should be.
* Problem Description: 
|{\color{blue}{\{problem\}}}|
* Test Case Input:
|{\color{blue}{\{input\}}}|
Please strictly follow the format below to return your answer:
Expected Output Format:
```json
|{\color{blue}{\{expected\_output\}}}|
```
Remember that your task is to predict the output of the test case, and you need to return the output in JSON format.
\end{minted}
\end{tcolorbox}

\nolinenumbers
\begin{lstlisting}[
  caption=An Example of the Reasoning Process for Test Case Generation,
  label={lst:reasoning_for_test_case},
  basicstyle=\ttfamily\scriptsize,
  breaklines=true,
  breakatwhitespace=true,
  columns=fullflexible,
  keepspaces=true,
  showstringspaces=false,
  frame=single,
  framerule=0.3pt,
  framesep=4pt,
  xleftmargin=0.01\linewidth,
  xrightmargin=0.01\linewidth,
  aboveskip=6pt,
  belowskip=6pt
]

Assistant: Okay, I need to create a test case for this problem. Let me first understand the problem thoroughly.

The problem is about a game between Takahashi and Aoki using cards. They take turns playing cards from their hands onto the table, and during their turn, they can also take a card from the table if it's smaller than the card they just played. The player who can't make a move loses.

The task is to determine who wins if both play optimally.

The constraints are that N + M + L <= 12, which is a small number, so the solution can afford to explore all possible states without performance issues.

The provided code uses memoization with functools.lru_cache to cache the results of the game states. It represents the state as the remaining cards in Takahashi's hand, Aoki's hand, the sorted table cards, and whose turn it is.

Now, I need to generate a test case that's valid, meaning the inputs must satisfy the constraints: N, M, L are at least 1, and their sum is <=12. The test case should not be identical to any example provided.

Looking at the sample inputs, they all have N, M, L as 1, 1, 2 or similar. To create a new test case, I can choose different values.

Let me think of a case where the outcome is Takahashi wins. Maybe a simple scenario where Takahashi can always force a win.

For example, let's take N=2, M=1, L=3. A = [1, 2], B = [3], C = [4,5,6].

In this case, Takahashi plays 1, then Aoki plays 3, but the table has 1, so Aoki can take 1. Then Takahashi plays 2, and the table has 2, so he can take 1 again. Aoki has no moves left, so Takahashi wins.

Wait, but I need to ensure that the test case is valid and that the code can handle it correctly.

Alternatively, perhaps a case where the table initially has a card that can be taken by both players.

Wait, but the initial state includes the table cards. So, for example, if the table has a card that is smaller than both players' cards, it can affect the outcome.

Let me create a test case where the initial table has a card that allows Takahashi to take it on his first move, putting Aoki in a bad position.

For example:

N=1, M=1, L=1
A = [2]
B = [3]
C = [1]

So, Takahashi plays 2, then checks the table for cards <2. The table has 1, so he takes it. Now, his hand is empty, Aoki's hand is [3], and the table has 3.

Aoki's turn: he plays 3, checks the table for <3, which is none, so he can't take anything. Now, Takahashi has no moves because he has no cards left, so Aoki wins.

Wait, but in this case, Takahashi took the 1, so after his move, the table has 3. Aoki plays 3, no cards to take, so Aoki can't make a move? Wait, no, Aoki plays 3, and then checks if there's a card on the table less than 3. The table has 3, so no. So Aoki's move is valid, and after that, the table is empty. Then Takahashi has no cards left, so he can't play anything and loses.

So the output should be Aoki.

But I'm not sure if this test case is correctly capturing any specific scenario. Maybe I should create a test case where Takahashi can win.

Alternatively, let's consider a case where Takahashi has a higher card than all on the table, and Aoki can't respond.

For example:

N=1, M=1, L=1
A = [3]
B = [2]
C = [1]

Takahashi plays 3, takes 1 if possible. Then the table has 2. Aoki plays 2, takes 1 if possible. Then Takahashi has no cards left, so Aoki wins again.

Hmm, maybe it's tricky to find a test case where Takahashi wins. Perhaps the provided code handles all cases correctly, but I need to ensure that the test case is valid.

Wait, the sample input 1 has N=1, M=1, L=2, and the output is Aoki. So perhaps a test case where the outcome is Takahashi can be created.

Let me try:

N=1, M=1, L=1
A = [1]
B = [2]
C = [3]

Takahashi plays 1, checks the table for <1, which is none, so he can't take anything. Now, the table has 3. Aoki's turn: he plays 2, checks the table for <2, which is none, so he can't take anything. Now, Takahashi has no cards left, so he can't play, and Aoki wins. Output is Aoki.

Another test case:

N=2, M=1, L=2
A = [2, 3]
B = [1]
C = [4,5]

Takahashi can play 2, take 4 if possible? Wait, 4 is not less than 2, so he can't take it. So the table has 2 and 4 and 5. Aoki plays 1, takes 2 if possible. Since 2 >1, he can't take it. So Aoki's move is valid, and the table has 4 and 5. Then Takahashi plays 3, checks for <3, which is none, so he can't take anything. Aoki has no cards left, so he can't play and loses. Takahashi wins.

Wait, but in this case, after Takahashi plays 2, the table has 2,4,5. Aoki plays 1, and since 1 <2, he can take 2. So Aoki takes 2, adds it to his hand, and the table now has 4 and 5.

Then Takahashi's turn: he has 3 left. He plays 3, checks the table for <3, which is none, so he can't take anything. Aoki has no cards left, so he can't play and loses. So Takahashi wins.

So the test case would be:

Input:
2 1 2
2 3
1
4 5

Output: Takahashi

This should be a valid test case, and it's different from the samples provided.

So, the test case would be:

N=2, M=1, L=2
A = [2,3]
B = [1]
C = [4,5]

The expected output is Takahashi.

This test case should be valid, and it's different from the samples, so it's a good candidate to test the code.
</think>

```json
[
  {
    "input": "2 1 2\n2 3\n1\n4 5",
    "output": "Takahashi"
  }
]
```

\end{lstlisting}
\linenumbers

\end{document}